\documentclass[10pt,twocolumn,letterpaper]{article}
\usepackage{iccv}
\usepackage{times}
\usepackage[numbers]{natbib}
\usepackage{amsmath}
\usepackage{amssymb}
\usepackage{booktabs}
\usepackage{caption}
\usepackage{cuted}
\usepackage{graphicx}
\usepackage{url}
\usepackage{xcolor}
\usepackage[pagebackref=true,breaklinks=true,colorlinks,bookmarks=false]{hyperref}
\usepackage{cleveref}

\newif\ifdark
\IfFileExists{/Users/vedaldi/.bash_profile}{
\immediate\write18{%
if defaults read -g AppleInterfaceStyle 2>/dev/null;
then echo \\darktrue > /tmp/displaymode.tex;
else echo \\darkfalse > /tmp/displaymode.tex; fi}
\input{/tmp/displaymode.tex}
\ifdark
\definecolor{pcolor}{HTML}{1E1E1E}
\definecolor{tcolor}{HTML}{C5C5C5}
\else
\definecolor{pcolor}{HTML}{FDF6E3}
\definecolor{tcolor}{HTML}{333333}
\fi
\pagecolor{pcolor}
\color{tcolor}
\hbadness=\maxdimen
\vbadness=\maxdimen
\vfuzz=30pt
\hfuzz=30pt
}{} 

\graphicspath{{.},{./images/qualitative/}}

\newcommand{\method}{DP3D\xspace}
\newcommand{\J}{\mathbf{J}}
\newcommand{\V}{\mathbf{V}}
\newcommand{\W}{\mathbf{W}}
\newcommand{\X}{\mathbf{X}}
\newcommand{\Y}{\mathbf{Y}}
\newcommand{\Z}{\mathbf{Z}}

\makeatletter
\renewcommand{\paragraph}{%
  \@startsection{paragraph}{4}%
  {\z@}{0.5em}{-1em}%
  {\normalfont\normalsize\bfseries}%
}
\makeatother

\iccvfinalcopy
\def\arxivsubmission{}
\def\iccvPaperID{9147}
\ificcvfinal\fi

\title{DensePose 3D\@: Lifting Canonical Surface Maps of Articulated Objects to the Third Dimension}
\author{Roman Shapovalov \quad David Novotny \quad Benjamin Graham \quad Patrick Labatut \quad Andrea Vedaldi\\
Facebook AI Research\\
{\tt\small \{romansh, dnovotny, benjamingraham, plabatut, vedaldi\}@fb.com}
} 

\begin{document}
\maketitle
\begin{strip}
\vspace{-4em}
\begin{center}
\includegraphics[width=\linewidth]{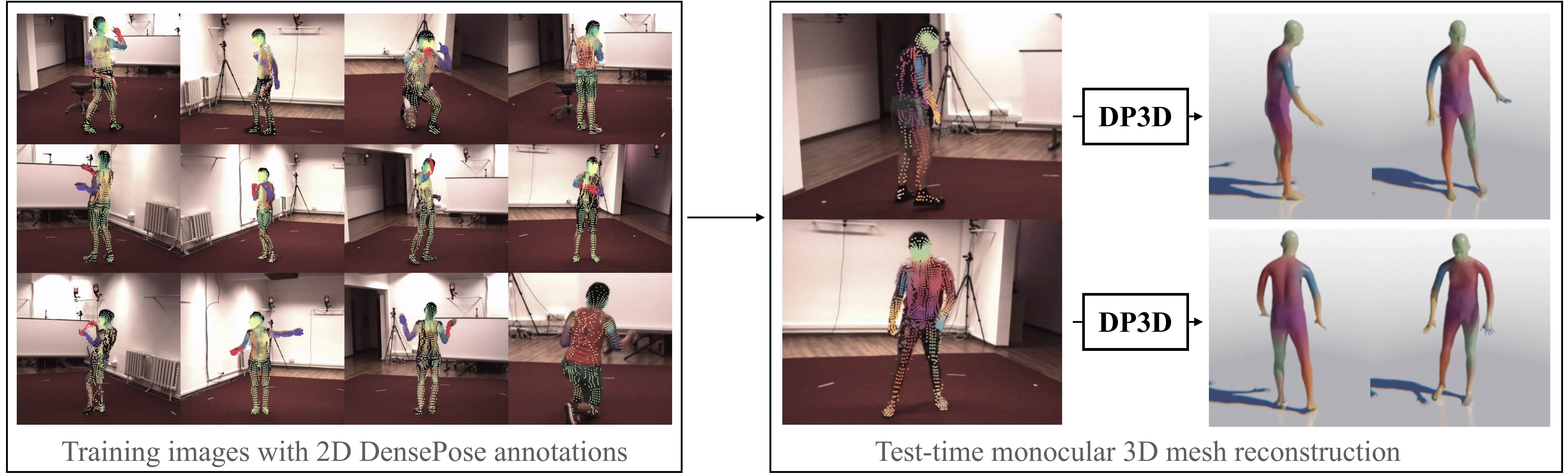}\\
\vspace{-0.7em}
\captionof{figure}{We propose DensePose 3D (DP3D), a method for monocular mesh recovery, which leverages a novel parametric mesh articulation model.
Crucially, the model is trained in a weakly supervised manner on a dataset of single views of humans or animals in different poses and their DensePose labelling produced by an off-the-shelf pre-trained detector.\vspace{-1em}%
}\label{f:splash}
\end{center}
\end{strip}

\begin{abstract}\vspace{-0.5em}
We tackle the problem of monocular 3D reconstruction of articulated objects like humans and animals.
We contribute DensePose~3D, a method that can learn such reconstructions in a weakly supervised fashion from 2D image annotations only.
This is in stark contrast with previous deformable reconstruction methods that use parametric models such as SMPL pre-trained on a large dataset of 3D object scans.
Because it does not require 3D scans, DensePose~3D can be used for learning a wide range of articulated categories such as different animal species.
The method learns, in an end-to-end fashion, a soft partition of a given category-specific 3D template mesh into rigid parts together with a monocular reconstruction network that predicts the part motions such that they reproject correctly onto 2D DensePose-like surface annotations of the object.
The decomposition of the object into parts is regularized by expressing part assignments as a combination of the smooth eigenfunctions of the Laplace-Beltrami operator.
We show significant improvements compared to state-of-the-art non-rigid structure-from-motion baselines on both synthetic and real data on categories of humans and animals.
\end{abstract}

\section{Introduction}

Recent advances in deep learning have produced impressive results in monocular 3D reconstruction of articulated and deformable objects, at least for particular object categories such as humans.
Unfortunately, while such techniques are general in principle, their success is rather difficult to replicate in other categories.
Before learning to reconstruct 3D objects from images, one must first learn a model of the possible 3D shapes of the objects.
For humans, examples of such models include SMPL~\cite{loper15smpl:} and GHUM~\cite{xu2020ghum}.
Constructing these requires a large dataset of 3D scans of the objects deforming and articulating over time, which have to be acquired with specialised devices such as domes.
Not only this hardware is uncommon, complex and expensive, but it is also difficult if not impossible to apply to many objects of interest, such as wild animals or even certain types of deformable inanimate objects.
Then, after building a suitable 3D shape model, one still has to train a deep neural network regressor that can predict the shape parameters given a 2D image of the object as input~\cite{kolotouros19convolutional,zanfir2020neural,Kanazawa2018}.
Supervising such a network requires in turn a dataset of images paired with the corresponding ground-truth 3D shape parameters.
Images with paired reconstructions are also very difficult to obtain in practice.
Some images may be available from the same scanners that have been used to construct the 3D model in the first place, but these are limited to `laboratory condition' by definition.
Thus, while there is abundance of `in the wild' images of diverse object categories that can be obtained from the Internet, they are lacking 3D ground-truth and are thus difficult to use for learning 3D shape predictors.

In this paper, we are interested in bootstrapping 3D models and monocular 3D predictors \emph{without} using images with corresponding 3D annotations or even unpaired 3D scans.
Fortunately, other modalities can provide strong cues for reconstruction.
For example, previous work~\cite{Kanazawa2018,novotny19c3dpo,kong2019deep,chen2019learning} leveraged 2D annotations for semantic keypoints to accurately reconstruct various object categories.
While these keypoints provide a supervisory signal at sparse image locations, DensePose~\cite{guler18densepose:,neverova2020continuous,sanakoyeu2020transferring} provides \emph{dense} correspondences between the images of humans or other animals and 3D templates of these categories.
Example of these annotations are shown on the left of~\Cref{f:splash}, where the colours encode the indices of corresponding points on the template mesh.
DensePose annotations can be seen as generalising sparse joint locations, with two important differences:
the density is much higher, and the correspondences are defined on the \emph{surface} of the object rather than in its skeleton joints.
Such dense annotations can be obtained manually or with detectors pre-trained on those manual 2D annotations, with the same degree of flexibility and generality as sparse 2D landmarks, while providing much stronger cues for learning detailed 3D models of the objects.
However, such annotations do not appear to have been used to bootstrap 3D object models before.

The main goal of this work is thus to leverage dense surface annotations, such as the ones provided by DensePose, in order to learn a parametric model of a 3D object category without using any 3D supervision.
As done in~\cite{Kanazawa2018,novotny19c3dpo,kong2019deep,chen2019learning}, we further aim to learn a deep neural network predictor that aligns the model to individual 2D input images containing the object of interest.
Our method assumes only having an initial \textit{rigid} canonical 3D template of the object category generated by a 3D artist.
There is no loss of generality here since knowledge of the template is required to collect DensePose annotations in the first place.%
\footnote{The 3D template is used by the human annotators as a reference to mark correspondences and defines the canonical surface mapping for the object category.}
Thus, pragmatically, we include this template in our model.

Our main contribution is a novel parametric mesh model for articulated object categories, which we call \emph{DensePose~3D} (DP3D).
In a purely data-driven manner, DP3D learns to softly assign the vertices of the initial rigid template to one of a number of latent parts, each of which moving in a rigid manner.
The parametrization of the mesh articulation is then given by a set of per-part rigid transforms expressed in the space of the logarithms of $\textsf{SE}(3)$.
In order to pose the mesh, each vertex of the template shape is deformed with a vertex-specific transformation defined as a convex combination of the part-specific transforms, where the weights are supplied by the soft segmentation of the corresponding vertex.
In order to prevent unrealistic shape deformations, we enforce smoothness of the part segmentation, and consequently of the vertex-specific offsets, by expressing the part assignment as a function of a truncated eigenbasis of the Laplace-Beltrami operator computed on the template mesh, which varies smoothly along the mesh surface.
We further regularise the mesh deformations with the as-rigid-as-possible (ARAP) soft constraint.

DP3D is trained in a weakly supervised manner, in the sense that our pipeline (including DensePose training) does not require 3D annotations for the input images.
In an end-to-end fashion, we train a deep pose regressor that, given a DensePose map extracted from an image, predicts the shape deformation parameters, poses the mesh accordingly, and minimises the distance between the projection of the posed mesh to the image plane and the input 2D DensePose annotations.
We show that our method does not need manual DensePose annotations for the training images; it can learn even from the predictions of a DensePose model trained on a different dataset.
This way, DP3D can learn to infer the shape of humans and animals from an unconstrained dataset containing diverse poses.
Since DP3D does not use images directly but only the DensePose annotations or predictions,
it is robust to changes in the object appearance statistics, which makes it suitable for transfer learning.

We conduct experiments on a synthetic dataset of human poses, and on the popular Human~3.6M benchmark, showing that the model trained on staged Human~3.6M generalises to a more natural 3DPW dataset.
We also fit the models to animal categories in the LVIS dataset.
Note that learning reconstruction of LVIS animals would be impossible with any method requiring 3D supervision since there are no scans or parametric models available for species like bears or zebras.
DP3D produces more accurate reconstructions than a state-of-the-art Non-rigid Structure-from-Motion (NR-SfM) baseline and compares favourably with fully-supervised approaches.

\section{Related work}\label{s:related}

In this section we review the relevant prior art: monocular human mesh reconstruction, canonical surface maps, and non-rigid SfM.


\paragraph{Image-based human body reconstruction.}

A popular method for reconstructing 3D humans from 2D images is test-time optimisation, where a parametric human model such as SMPL~\cite{loper15smpl:} or SCAPE~\cite{anguelov05scape:} is fitted to a given test image by minimising various types of energies, including 2D keypoint and mask reprojection losses~\cite{guan09estimating,sigal08combined,bogo16keep,Lassner2017,huang17towards,zanfir18monocular,joo18total,pavlakos19expressive,xiang19monocular}.
Alternatively, one can learn a deep regressor which, given a single image as input, predicts the parameters of the 3D shape model directly.
Most methods~\cite{anguelov05scape,mehta17vnect,rogez18lcr-net,sun18integral,kolotouros19convolutional, martinez17a-simple} reconstruct only a sparse set of 3D points, usually corresponding to 2D body joint detections.
HMR~\cite{kanazawa18end-to-end} and GraphCMR~\cite{kolotouros19convolutional} regress instead full 3D meshes.
Kolotouros~\etal~\cite{kolotouros19learning} combine the test-time optimization and deep regression paradigms.
Biggs~\etal~\cite{biggs20203d} regress multiple mesh hypotheses to deal with the inherent ambiguity of monocular 3D reconstruction.
While such methods achieve state-of-the-art monocular human mesh recovery, they require large dataset with 3D annotations to train the 3D shape model and the regressor.
In contrast, our method is trained only with 2D image annotations.

\paragraph{Self-supervised 3D human pose estimation.}

Other methods aim at reconstructing 3D body skeletons without 3D annotations.
Some works leverages multi-view constraints~\cite{kocabas2019self,pavlakos2017harvesting,rhodin2018learning} while Pavlakos \etal~\cite{pavlakos2018ordinal} assume ordinal depth supervision.
Alternatively, adversarial networks can also be used to learn 3D models from 2D annotations in a monocular setup~\cite{kudo2018unsupervised,drover20183dpose,chen2019unsupervised}.
The idea is to train a discriminator that tells if the 2D reprojection of the reconstructed 3D points from multiple random views is plausible or not.
While these methods work well, their inability to deal with occluded keypoints makes them unsuitable for dense reconstruction.

\begin{figure*}[tb]
\begin{center}
\includegraphics[width=\linewidth]{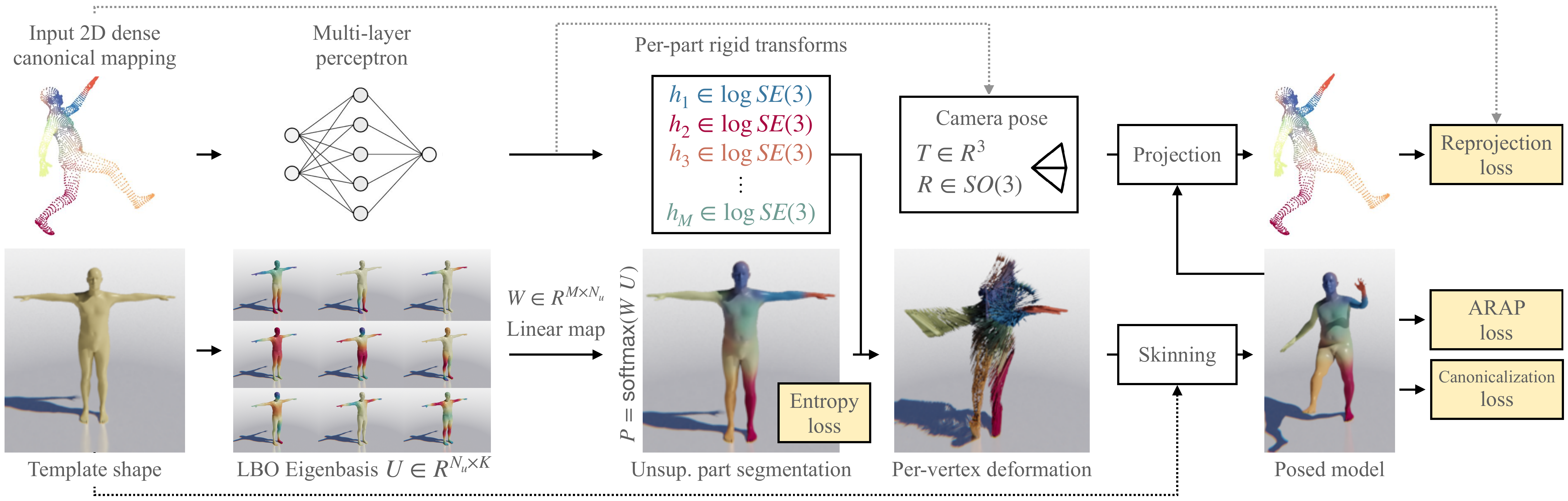}%
\vspace{-0.7em}%
\caption{\textbf{Overview of our method.}
The input 2D keypoints~$\Y$ are passed to the network~$\Phi$ that predicts global and per-part rigid transformations.
LBO harmonics are used to regress the soft part segmentation~$\mathbf{P}$.
The transformations, part segmentation, along with the template mesh~$\V$, are used for Linear Blend Skinning to obtain the shape~$\X$.
During training, this shape enters re-projection, canonicalisation, and ARAP losses, while the entropy loss is defined on part segmentation.%
\vspace{-0.5cm}%
}\label{f:overview}
\end{center}
\end{figure*}

\paragraph{Canonical surface maps.}

DensePose~\cite{guler18densepose:} was perhaps the first method to predict dense assignments from an image to a reference 3D template of the human body, also called a \emph{Canonical Surface Map} (CSM).
It introduced a dataset with manually labelled correspondences as well as a new deep network architecture to regress dense correspondences from images.
Follow-up work introduced semi-supervised learning~\cite{Neverova2019} and transferred human correspondences to quadrupeds~\cite{sanakoyeu2020transferring}.
Most recently, Neverova~\etal~\cite{neverova2020continuous} reformulated DensePose as a non-parametric problem by predicting canonical point embeddings for image pixels,
which facilitates its application to a wider range of deformable object categories.

Other works aimed at learning CSMs with limited or no supervision:
{}\cite{thewlis17dense,thewlis19unsupervised,schmidt17self-supervised} do so by using principles such as transformation equivariance, whereas~\cite{kulkarni19canonical} enforces consistency with an initial 3D model of the object.
Relevant to our work, the articulation-aware variant of it~\cite{Kulkarni2020} produces canonical surface maps for categories such as quadruped animals.
The method requires a segmented template mesh with a predefined skeleton structure; in contrast, we learn the articulated structure automatically without supervision.

\paragraph{Non-rigid structure-from-motion.}

NR-SfM is relevant to our work as its goal is to reconstruct a deformable 3D object from 2D keypoint annotations.
The seminal work of Bregler~\cite{bregler00recovering}, which proposed to express the possible deformations of the 3D shape as a linear combination of a small number of basis shapes, has since inspired many follow-up works~\cite{agudo2018image, fragkiadaki2014grouping, dai2014simple, zhu2014complex, akhter2009nonrigid, akhter2011trajectory, agudo2017dust, gotardo2011non, kumar2018scalable, kumar2017spatial, zhu2014complex, agudo2018deformable, zhou2016sparseness, zhou2016sparse, torresani2008nonrigid}.
Traditionally, such methods posed the problem as matrix factorization, but more recently some alternative that leverage deep learning have emerged.
DeepNRSfM~\cite{kong2019deep, wang2020deep} and, more relevantly to our work, C3DPO~\cite{novotny19c3dpo} train an MLP that maps the vectorised list of 2D keypoints to camera and shape parameters and minimise the distance between the input 2D keypoints and the 3D point reprojections.
While C3DPO works well with sparse keypoints such as the human joints, as we show in the experiments, it fails to handle the dense collections of points required to reconstruct meshes.
We address this issue by utilising the known category-level template mesh to learn deformations compatible with the articulation of a latent skeletal structure.

\section{Method}\label{s:method}

We aim to learn reconstructing the 3D shape of a deformable object such as a human or an animal from 2D images, and to do so without 3D supervision.
Instead, we only use dense 2D object points that can be annotated manually or predicted by means of a method such as DensePose, also known as a canonical surface map (CSMs).

We summarise the necessary CSM background in~\cref{s:csm} and then discuss our method.


\subsection{Canonical surface maps}\label{s:csm}

A CSM~\cite{thewlis17dense,guler18densepose:,Neverova2019,neverova2020continuous,kulkarni19canonical,sanakoyeu2020transferring,Kulkarni2020} is defined with respect to a reference 3D template, usually given as a triangular mesh with vertices
$
\V = (V_k)_{k = 1}^{K} \in \mathbb{R}^{K\times 3}.
$
For humans, for example, a common reference mesh is the SMPL rest pose (which was created by a 3D artist).

A CSM such as DensePose takes as input an image $I : \Omega \rightarrow \mathbb{R}$ of the object and assigns to each pixel $y \in \Omega$ a point in the mesh $\V$, producing a map $\Omega \rightarrow \V$.\footnote{In practice, the map is valued in $\V \cup \{\text{bkg}\}$ to allow to mark pixels that do not belong to the object as background.}
%
While this is useful information, it is not yet a 3D reconstruction of the object in the image because $\V$ is a fixed reference template.
In order to obtain a 3D reconstruction, we need instead to \emph{pose} the template by finding a suitable deformation
$
 \X = (X_k)_{k=1}^K \in \mathbb{R}^{K\times 3}
$
of its vertices.


As the first step in the posing process, we `reverse' the CSM output and, for each vertex $V_k$ of the template, find its corresponding pixel location $y_k$, resulting in a collection of 2D vertex locations
$
\Y = (y_k)_{k = 1}^{K} \in \mathbb{R}^{K\times 2}
$.
Due to occlusions, a vertex may be invisible in the image, which prevents extracting its 2D location~$y_k$ from the CSM\@.
Thus we also define visibility indicators
$
\Z = (z_k)_{k = 1}^{K} \in \{0, 1\}^{K}
$.

Note that $\Y$ can also be obtained from the posed mesh $\X$ and the camera projection function $\pi_I$ as
$
y_k = \pi_I(X_k).
$
This calculation does not involve the CSM at all and, as we show later, can be used to constrain the reconstruction.

\subsection{Shape model}\label{s:model}

In order to reconstruct the 3D shape of an object from 2D annotations, we must define a \emph{shape model} that constrains the space of possible reconstructions $\X$.
To this end, we assume that the underlying object, which could be a human or another animal, has a skeletal structure.
Under this assumption, the pose of the object is expressed by the rigid transformations of $M$ parts
\begin{equation}
g_m = (R_m,T_m) \in SE(3), \quad m=1,\dots,M.
\end{equation}
We assume that each vertex $\V_k$ in the template belongs to one of the $M$ parts with membership strength
$
P_{km} \in [0,1]
$
such that
$
\sum_{m=1}^M P_{km} = 1.
$
The posed vertices~$\X$ are given by the linear combination of part transformations, as in linear blend skinning (LBS):
\begin{equation}\label{e:skinning}
X_k = \sum_{m=1}^M P_{km} \cdot g_{m}(g_{0m}^{-1}(V_k)).
\end{equation}
Here $g_{0m} \in SE(3)$ stands for the rest pose of the $m$-th part.

While we do not force the parts to have a particular semantic, we expect learning to group together surface points that move rigidly together, e.g all points on a forearm.
Next, we explain how we encourage such a solution to emerge.

\paragraph{Part segmentation.}

Having defined per-vertex deformations, we will now describe the part segmentation model $\mathbf{P} = [P_{km}]\in\mathbb{R}^{K\times M}$.
As mentioned before, unlike other parametric models~\cite{loper15smpl:,anguelov05scape:}, we do not require a pre-segmented template shape.
Instead, we treat the part segmentation~$\mathbf{P}$ as a latent variable and learn it together with the rest of the model parameters.
Note that the part segmentation is independent of a particular input instance --- this means that the part assignments stay constant once training finishes.
Intuitively, limiting the number of parts and constraining deformations within parts to rigid ones should force the model to group the vertices that move according to the same rigid transform into the same part.

\paragraph{Smooth segmentation with LBO.}

While we have reduced the deformation of the template to the rigid motions of a small number of parts ($M=10$), the assignment of the template vertices to the different parts can still be irregular, which may lead to unrealistic body deformations.
We address this issue by enforcing the part assignments $\mathbf{P}$ to be smooth.
Combined with~\cref{e:skinning}, this encourages the deformations of the template to be smooth as well.

\begin{figure}
\newcommand{\lbofig}[1]{%
\includegraphics[trim=1cm 1cm 1cm 5cm, clip, width=0.24\linewidth]{images/lbo/lbo_#1_000_-35.png}%
}%
\centering%
\lbofig{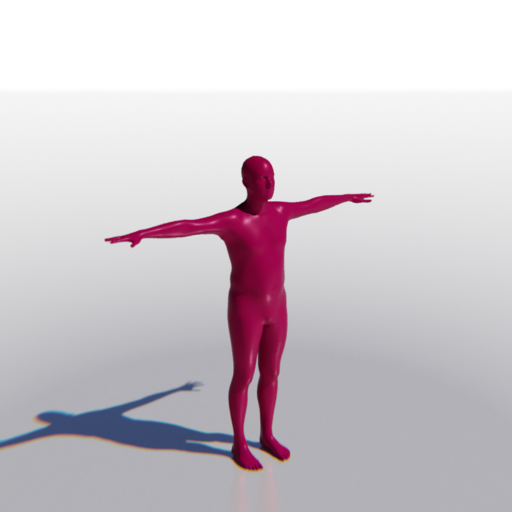}%
\lbofig{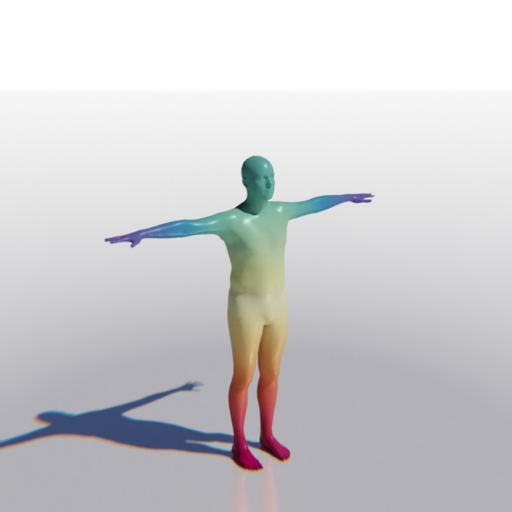}%
\lbofig{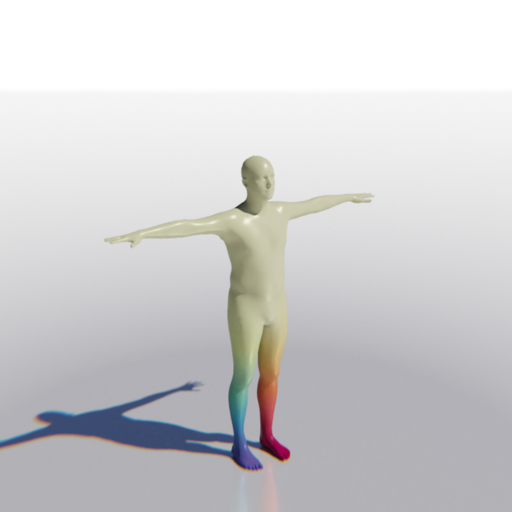}%
\lbofig{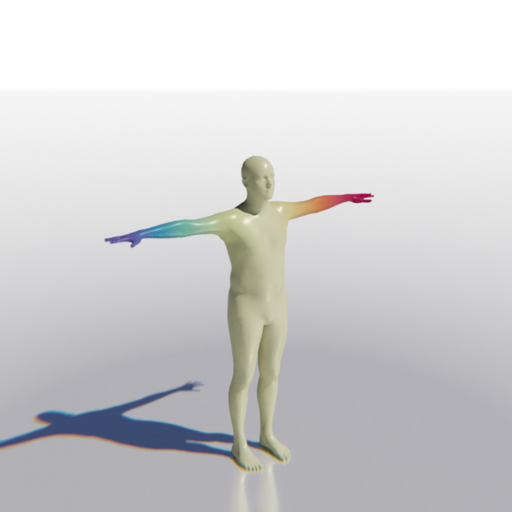}\\
\lbofig{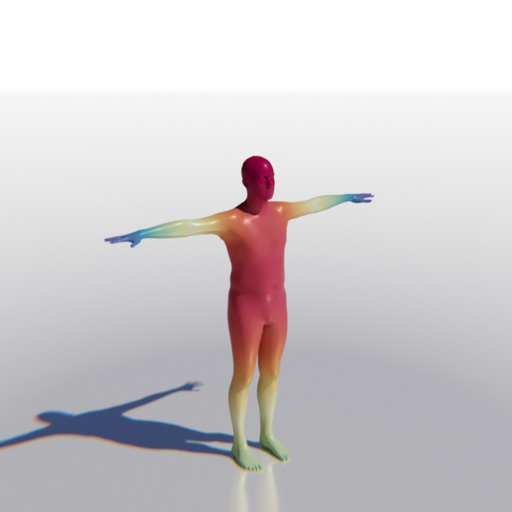}%
\lbofig{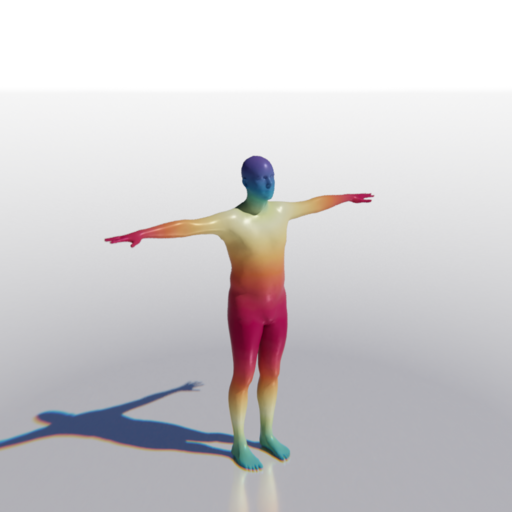}%
\lbofig{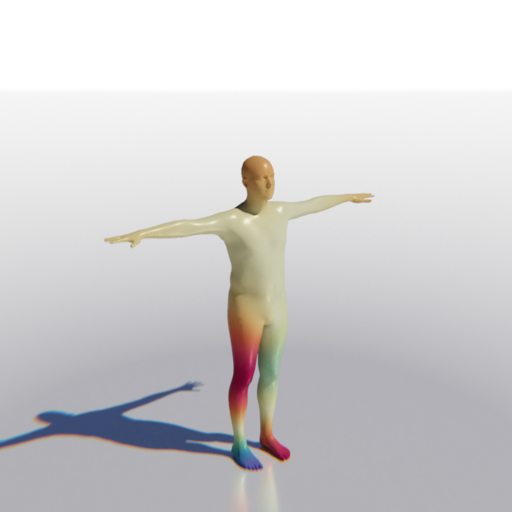}%
\lbofig{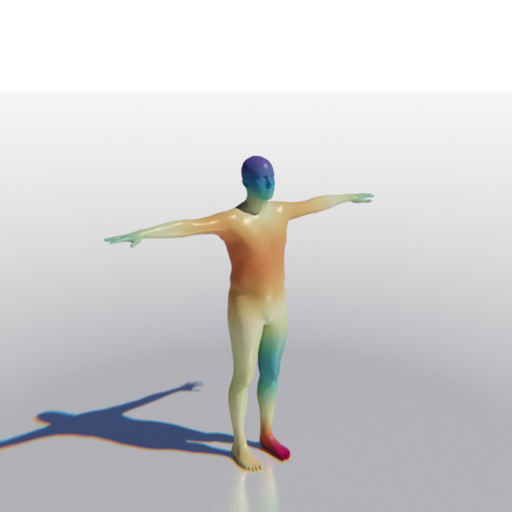}\\
\lbofig{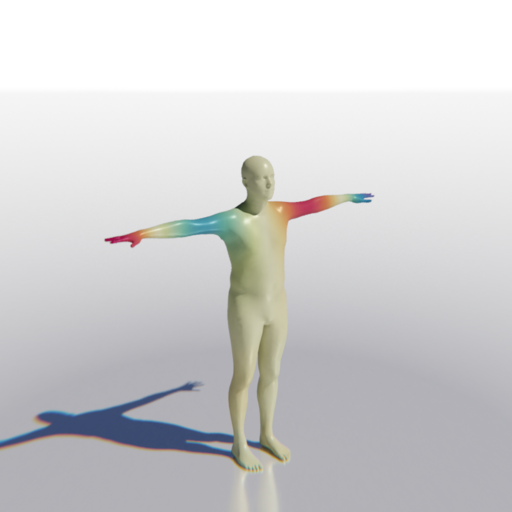}%
\lbofig{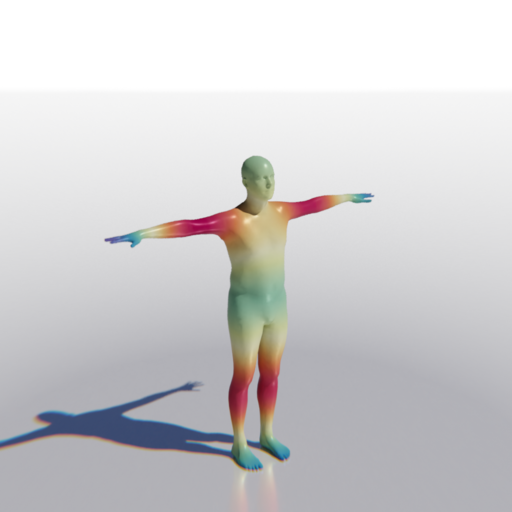}%
\lbofig{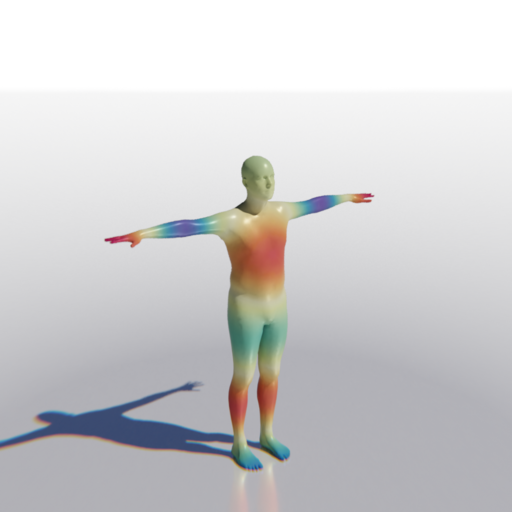}%
\lbofig{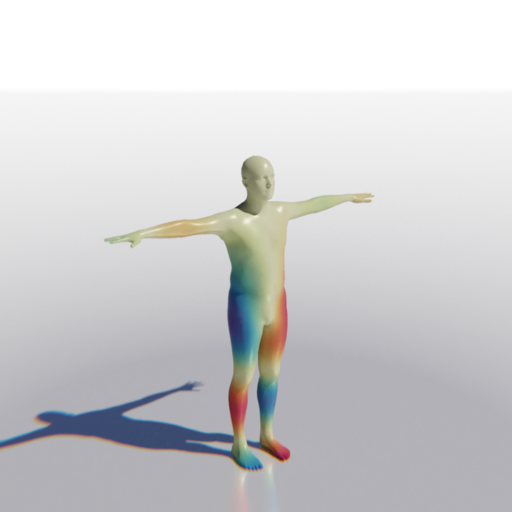}\\
\caption{We express the per-vertex deformations as a linear map of the Laplace--Beltrami eigenbasis of the template shape.
The figure shows 12 most significant eigenvectors of the LB operator of a human template mesh.
These top eigenfunctions vary smoothly along the surface of the mesh, 
which enforces similarity of the neighboring per-vertex transforms, leading to natural mesh deformations.}
\label{fig:lbo-fig}
\end{figure}

We formalise this intuition by requiring the part assignment~$\mathbf{P}$ to be a smooth function on the mesh surface.
This can be enforced by making sure that $\mathbf{P}$ only contains `low frequency' components.
Formally, this is achieved by expressing $\mathbf{P}$ as a linear combination of selected eigenfunctions of the Laplace-Beltrami operator (LBO~\cite{rustamov2007laplace}), illustrated in \Cref{fig:lbo-fig}.

In more detail, consider the discrete approximation $\Delta$ of the LBO for the reference template mesh $\V$.
Let $\mathbf{u}_i \in \mathbb{R}^{K}$ be the (orthonormal) eigenvectors of $\Delta$ sorted by increasing eigenvalue magnitude, and let $\mathbf{U} = (\mathbf{u}_i)_{i=1}^{N_u}\in\mathbb{R}^{K\times N_u}$ be the matrix containing the $N_u$ first eigenvectors.
We define the part segmentation as
\begin{equation}\label{e:parts}
\mathbf{P} = \textrm{softmax}(\mathbf{U} \mathbf{W}),
\end{equation}
where $\mathbf{W} \in \mathbb{R}^{N_u \times M}$ is a parameter matrix, and the softmax is taken with respect to the part index $k$.

Smoothness can be further increased by reducing~$N_u$ or by initialising~$\mathbf{W} = [W_{im}]$ with decreasing magnitude.
Specifically, we use a variant of Xavier initialisation and set
$
W_{im} \thicksim \mathcal{N}(0, \frac{\exp(- i / \bar{\sigma})}{M^{1/2}}).
$
This focuses the model on low-frequency harmonics at the beginning of training.

\paragraph{Transformations predictor.}

Given input 2D keypoint locations~$\Y$ and visibilities~$\Z$, we train a multi-layer perceptron (MLP) to predict the $(M+1)$ rigid part transformations:
\begin{equation}\label{e:transforms}
  \{h_m\}_{m=0}^M = \Phi(\Y, \Z).
\end{equation}
We express transformations in log-space, meaning that
$
(R_m, T_m) = g_m = \exp(h_m)
$
where~$\exp: \mathbb{R}^6 \to \mathbb{R}^{3 \times 3} \times \mathbb{R}^3$ is the exponential map of $SE(3)$; see~\cite{Blanco2010} for details.

Note that we estimate the additional global transformation $h_0$; this is the camera pose used to re-project the posed shape, which is expressed in the object reference frame, back to the image (see \cref{e:rep}).
Note also that \cref{e:skinning} requires the inverse part transformation at rest $g_{0m}^{-1}$; these are learnt as logarithms of canonical pose angles~$\mathbf{w}^r_0 \in \mathbb{R}^{M \times 6}$ so that~%
$
\forall m: \quad g_{0m}^{-1} = \exp(\mathbf{w}^r_{0m})
$.

\subsection{Training}

We train the MLP~\eqref{e:transforms}, mapping the 2D points to the pose parameters, and the part segmentation model~\eqref{e:parts} by combining a number of losses.


\paragraph{Re-projection loss.}

The first loss ensures that the posed mesh reprojects correctly onto the 2D points:
\begin{equation}\label{e:rep}
\mathcal{L}_\text{rep}
=
\frac{\sum_{k=1}^K 
z_k a_k
\left \|
y_k - \pi\left(X_k R_0 + T_0 \right)
\right \|}
{\sum_{k=1}^K z_k a_k},
\end{equation}
where $X_k$ and $(R_0,T_0)$ are obtained by composing the pose regressor~\eqref{e:transforms} with the skinning function~\eqref{e:skinning}.
We weigh the mesh vertices~$V_k$ with the areas~$a_k$ of the corresponding barycells to make the loss resampling-invariant.

\paragraph{Canonicalisation loss.}

The authors of C3DPO~\cite{novotny19c3dpo} proposed the \emph{canonicalisation} loss to remove the ambiguity in recovering the camera pose and a 3D reconstruction, which also helps with overfitting.
The idea is to learn an auxiliary network $\hat \X \approx \Psi(\hat{\X}\tilde{R})$ tasked with undoing a random rotation $\tilde{R}$ applied to the point cloud $\hat\X$ (defined in the object coordinates).
Novotny~\etal~\cite{novotny19c3dpo} prove that this loss can be minimised only if the predicted shapes~$\hat\X$ are indeed canonical w.r.t.~orientation, meaning that the model cannot predict two different reconstructions $(\hat\X_1, \hat\X_2)$ that only differ by a rigid transformation.
Specifically, the loss is formulated as
\begin{equation}\label{e:canon}
\mathcal{L}_\text{canon}
=
\sum_{k=1}^K 
\left \|
\left[\hat{\X} - \Psi(\hat{\X} \tilde{R}) \right]_k
\right \|,
\end{equation}
where~$\tilde{R} \in \mathbb{R}^{3 \times 3}$ is a random rotation matrix and $[\cdot]_k$ extracts the $k$-th row of its argument.


\paragraph{ARAP loss.}

To further increase the robustness of the reconstruction, we encourage the deformation of the template shape to be as-rigid-as-possible (ARAP)~\cite{Sorkine2007}.
This is particularly useful when, as it is often the case, the input DensePose annotations are noisy and biased.
ARAP measures the cost of deforming the template mesh $\V$ into the posed mesh $\X$:
\begin{equation}\label{e:arap}
\mathcal{L}_\text{arap}(\X;\V)
=
\sum_{k=1}^K
\min_{R \in \textit{SO(3)}}
\sum_{q \in \mathcal{N}(k)} 
w_{kq} 
\left \|
V_{\vec{kq}} - X_{\vec{kq}}R
\right \|,
\end{equation}
where $\mathcal{N}(k)$ denotes indices of adjacent template vertices,
$V_{\vec{kq}} = V_q - V_k$, $X_{\vec{kq}} = X_q - X_k$, 
and weights~$w_{kq}$ are defined proportionally to the area of the faces incident to the edge~$kq$;
see~\cite{Sorkine2007} for details.
We back-propagate the error through estimated coordinates~$X_k$ and~$X_q$ but stop gradients after fitting the rotation~$R$.

\paragraph{Entropy regularisation.}

Sometimes the model tends to assign several part indices to a single vertex, which makes deformations too rigid.
We thus regularise the segmentation model by penalising the entropy of the part distribution for each vertex using the following loss:
\begin{equation}\label{e:entropy}
\mathcal{L}_{\textrm{entropy}} = - \frac{1}{K} \sum_{k=1}^K \sum_{m=1}^M P_{km} \log P_{km}.
\end{equation}

\paragraph{Learning formulation.}

To train the method, we optimise the parameters of the networks $\Phi$, $\Psi$, and the matrix~$\W$ (\cref{e:parts}) minimising a weighed combination of the losses above:
\begin{equation}\label{e:totalloss}
\mathcal{L}
=
\mathcal{L}_{\textrm{rep}} +
w_{\textrm{entropy}} \mathcal{L}_{\textrm{entropy}} +
w_{\textrm{canon}} \mathcal{L}_{\textrm{canon}} +
w_{\textrm{arap}} \mathcal{L}_{\textrm{arap}}.
\end{equation}
Loss weights~$\mathbf{w}$ are treated as hyper-parameters, see supplementary material for the values used for the experiments.
\section{Experiments}\label{s:experiments}

\graphicspath{{images/qualitative/}}
\newcommand{\cdpodir}{dpc3dpo-up3d_026_c3dpo_00000_M.l.l_canonicalization=0.1_MOD.camera_scale=True_MOD.argmin_translation=True}
\newcommand{\oursdir}{dpc3dpo-up3d_040_ablation_00015_M.l.l_canonicalization=0.1_M.l.l_part_entropy=0.001_M.l.l_local_rigidity=0.0_M.l.l_arap=0.3_MOD.shape_basis_size=0}

\newcommand{\imh}{1.8cm}

\newcommand{\methodfig}[2]{%
\includegraphics[trim=2cm 1cm 2cm 2cm, clip, height=\imh]{#1/#2.png}%
}

\newcommand{\imfig}[1]{%
\includegraphics[height=\imh]{\oursdir/#1_image_kp_lbo.png}%
}

\newcommand{\rainbowours}{%
{\color[HTML]{8D274B}D}%
{\color[HTML]{A48053}P}%
{\color[HTML]{547CA2}3}%
{\color[HTML]{8EA395}D}%
}

\newcommand{\methtext}[1]{\centering\scriptsize\hspace{0.01cm}\rotatebox{90}{\hspace{0.01cm}#1}\hspace{0.01cm}}

\newcommand{\comparerow}[1]{%
\begin{minipage}[c]{0.24\linewidth}\centering%
    \imfig{#1}\\
    \methtext{~~~~\textbf{\color[HTML]{960061}C3DPO \cite{novotny19c3dpo}}}%
    \methodfig{\cdpodir}{#1_pcl_raw_solid_000_000}%
    \methodfig{\cdpodir}{#1_pcl_raw_solid_000_089}\\
    \methtext{~~~~\textbf{\rainbowours~\color[HTML]{000000}(ours)}}%
    \methodfig{\oursdir}{#1_mesh_raw_partseg_000_000}%
    \methodfig{\oursdir}{#1_mesh_raw_partseg_000_089}%
\end{minipage}}


\begin{figure*}
\comparerow{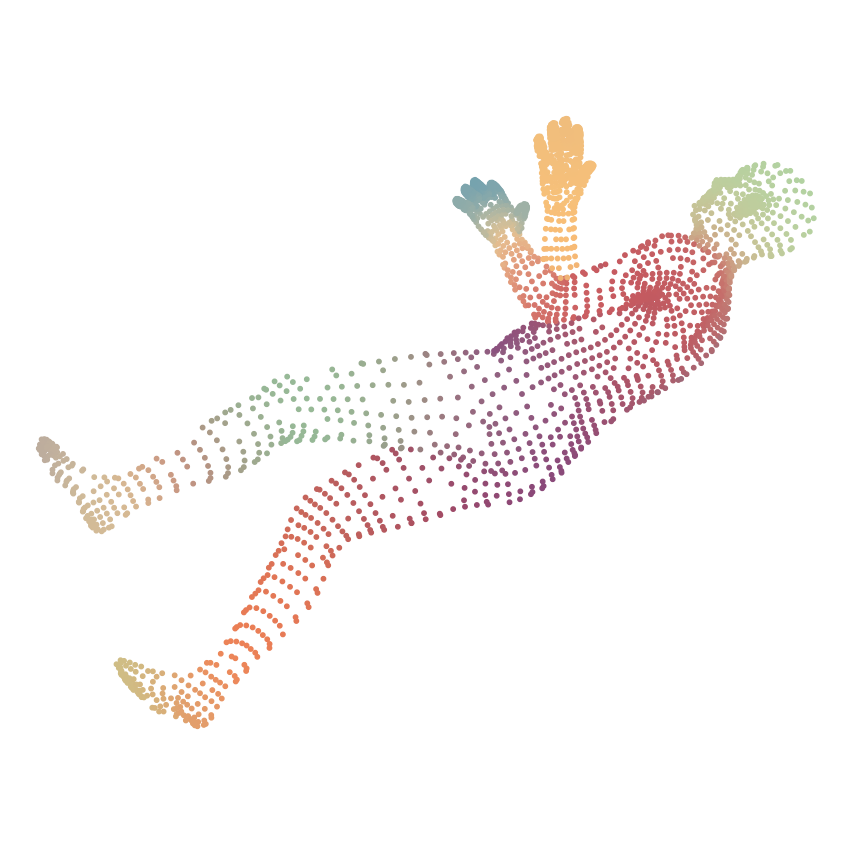}
\comparerow{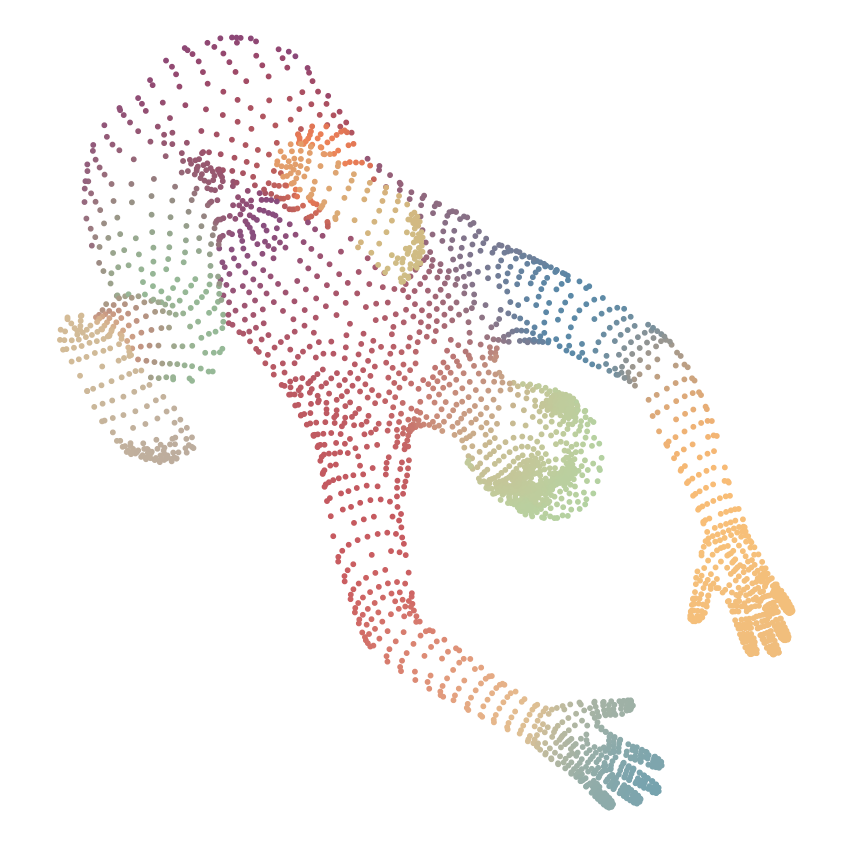}
\comparerow{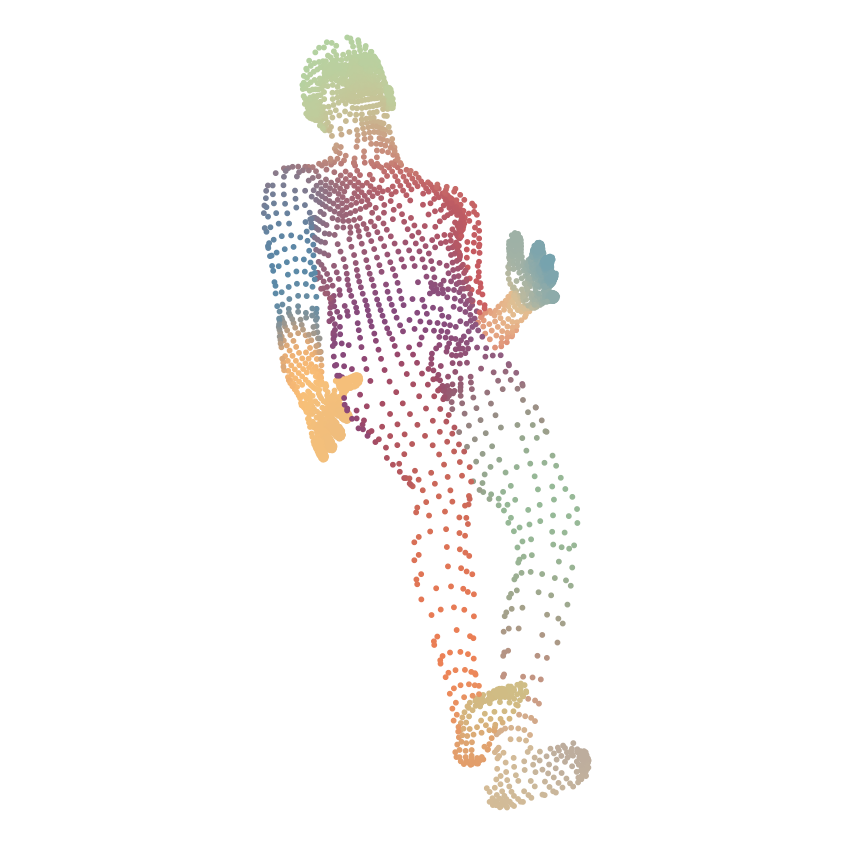}
\comparerow{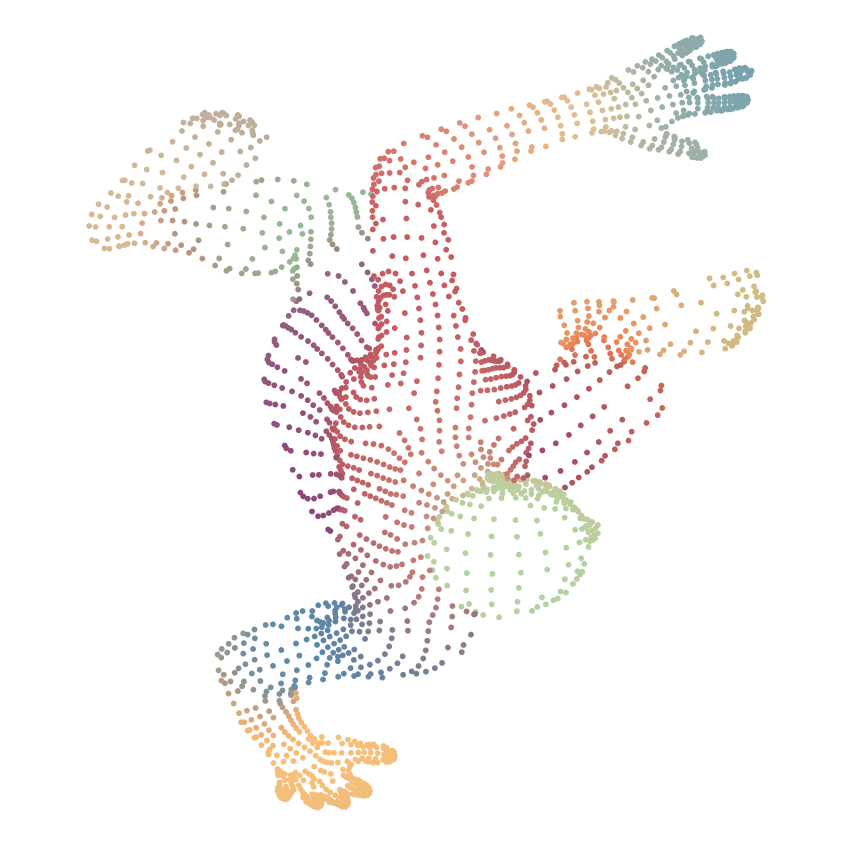}\\

\caption{\textbf{Qualitative comparison on UP-3D}. The figure shows the input keypoints (colours encode keypoint indices),
the reconstruction of C3DPO~\cite{novotny19c3dpo},
and of our method (DP3D), where colours correspond to the learnt part segmentation.%
}\label{f:up3dres}
\end{figure*}

\let\cdpodir\undefined
\let\oursdir\undefined
\let\imh\undefined
\let\cdpodir\undefined
\let\oursdir\undefined
\let\imh\undefined
\let\methodfig\undefined
\let\imfig\undefined
\let\rainbowours\undefined
\let\methtext\undefined
\let\comparerow\undefined

We evaluate the quality of our reconstruction on human and animal data, both synthetic and real, and then ablate various components.
We compare our results to C3DPO~\cite{novotny19c3dpo} because it is the best-performing Non-Rigid SFM approach that works under assumptions compatible with ours.

Implementation details of the networks and training are provided in the sup.~mat.
We will share the Pytorch code.

\subsection{Datasets and metrics}

\paragraph{UP-3D and Stanford Dogs.}

First, we evaluate the method on two clean, synthetic datasets: UP-3D (humans) and Stanford Dogs.
UP-3D~\cite{Lassner2017} contains SMPL fits for 8515 photos of people rendered under 30 random viewpoints.
We orthographically project the mesh vertices and input their ground-truth visibility and vertex identity to \method directly (instead of using DensePose for UV extraction).
For Stanford Dogs, we follow UP-3D and fit a dog model to a subset of ImageNet using SMALify~\cite{biggs18creatures} on the mask and 2D keypoint annotations provided in StanfordExtra dataset~\cite{biggs2020wldo}.
We obtain in this way 6511 training and 4673 test instances.
Please refer to the sup.~mat.~for further details.

We report the mean per-joint position error (MPJPE) of the reconstructions.
Since we have the ``ground-truth'' SMPL/SMAL model~$\X$ for each test instance, we compute MPJPE of the estimated shape in camera coordinates~$\bar\X = \hat\X R_0 + T_0$ using all keypoints (not only the visible ones):
$
\textrm{MPJPE}(\bar\X, \X) = \frac{1}{K} \sum_{k=1}^K \|\bar{X_k}- X_k \|.
$
Since via orthographic projection depth is only known up to a constant, we normalise depth by subtracting its mean from $\X$ and~$\bar\X$ before computing the loss.
We use the original train/test splits.

\graphicspath{{images/qualitative/}}
\newcommand{\lbocdpodir}{dpc3dpo-standogs_007_lboc3dpo_00005_MOD.shared_scale=True_M.l.l_canonicalization=0.1_MOD.lbo_num_harmonics=64_MOD.weight_reprojection_by_area=True}
\newcommand{\oursdir}{dpc3dpo-standogs_010_ablation_00033_MOD.shared_scale=True_M.l.l_canonicalization=0.1_M.l.l_part_entropy=0.001_MOD.lbo_num_harmonics=64_MOD.num_latent_parts=10_MOD.shape_basis_size=5}

\newcommand{\imh}{1.8cm}

\newcommand{\methodfig}[2]{%
\includegraphics[trim=1cm 0.5cm 1cm 1cm, clip, height=\imh]{#1/val_#2.png}%
}

\newcommand{\imfig}[1]{%
\includegraphics[height=\imh]{\oursdir/val_#1_image_kp_partseg}%
}

\newcommand{\rainbowours}{%
{\color[HTML]{8D274B}D}%
{\color[HTML]{A48053}P}%
{\color[HTML]{547CA2}3}%
{\color[HTML]{8EA395}D}%
}

\newcommand{\methtext}[1]{\centering\scriptsize\hspace{0.01cm}\rotatebox{90}{\hspace{0.01cm}#1}\hspace{0.01cm}}

\newcommand{\comparerow}[1]{%
\begin{minipage}[c]{0.24\linewidth}\centering%
    \imfig{#1}\\
    \methtext{~~~~w/o parts}%
    \methodfig{\lbocdpodir}{#1_mesh_raw_solid_000_000}%
    \methodfig{\lbocdpodir}{#1_mesh_raw_solid_000_089}\\%
    \methtext{~~~~\textbf{\rainbowours~\color[HTML]{000000}(ours)}}%
    \methodfig{\oursdir}{#1_mesh_raw_partseg_000_000}%
    \methodfig{\oursdir}{#1_mesh_raw_partseg_000_089}\\%
    \methtext{~~GT SMAL fits}
    \methodfig{\oursdir}{#1_gt_mesh_raw_err_000_000}
    \methodfig{\oursdir}{#1_gt_mesh_raw_err_000_089}
\end{minipage}}

\begin{figure*}
\comparerow{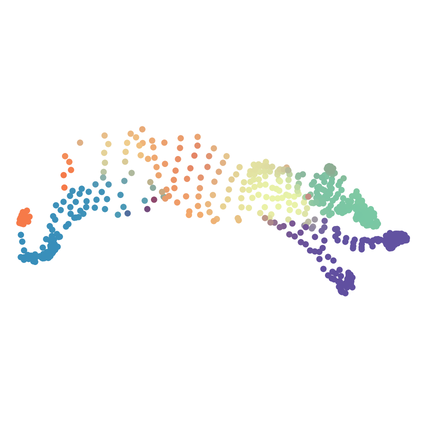}
\comparerow{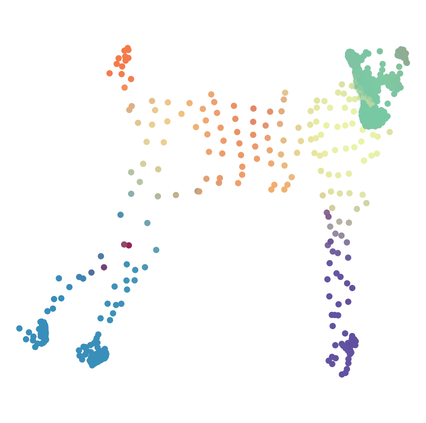}
\comparerow{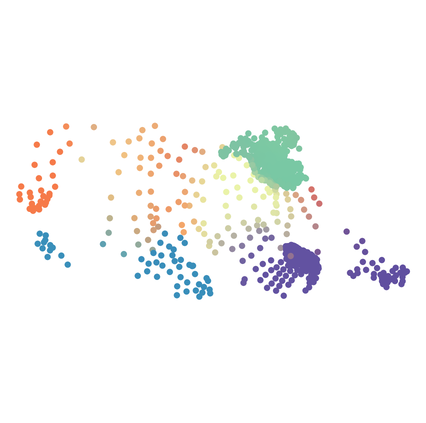}
\comparerow{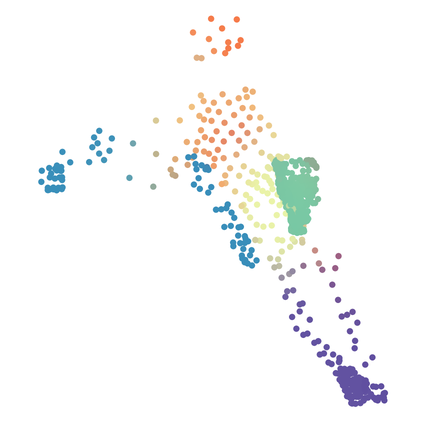}

\caption{\textbf{Results on Stanford Dogs}. 
The first row shows input keypoints obtained by projecting SMAL fits from the last row,
the middle rows show the result of the no-parts baseline and of our reconstruction from the camera's and from an alternative viewpoint (colours correspond to the learnt part segmentation),
the last row color-codes errors on the ``ground-truth'' mesh.%
\vspace{-0.5cm}%
}\label{f:dogsres}
\end{figure*}

\let\cdpodir\undefined
\let\lbocdpodir\undefined
\let\oursdir\undefined
\let\imh\undefined
\let\cdpodir\undefined
\let\oursdir\undefined
\let\imh\undefined
\let\methodfig\undefined
\let\imfig\undefined
\let\rainbowours\undefined
\let\methtext\undefined
\let\comparerow\undefined

\paragraph{Human 3.6M} consists of real images of 7 people equipped with motion-capture sensors performing various tasks in the lab environment.
The dataset provides the locations of 3D joints rather than full body surface.
Hence, for evaluation purposes, we compute the mean reconstruction error on $N_J = 14$ joints (RE$_{14}$).
In order to obtain the joints' positions~$\hat\J$ from the posed mesh  $\hat\X$, we run the pre-trained linear joint regressor from the SMPL model~\cite{loper15smpl:}: if the resulting joints are correct, the mesh must have been posed correctly.
We rigidly align the sets of points and find the optimal scale before computing the metric:
$
\textrm{RE}(\hat\J, \J) = \min_{s,R,T} \frac{1}{N_J} \sum_{i=1}^{N_J} \|(s\hat{J_i}R + T) - J_i \|.
$

For training, we sampled the videos at 10 frames per second, resulting in 311,424 images.
For evaluation, we use the scheme known as `Protocol \#1'.
The test set videos are sampled at 25 FPS, resulting in 109,792 images.
We ran the pre-trained DensePose detector from Detectron2~\cite{wu2019detectron2} on all images independently to obtain the input UV annotation,
then converted them to 2D projections of SMPL vertices as described in~\Cref{s:csm}.
We use the standard train/test split, setting out all images of subjects 9 and 11 for testing.

\paragraph{3DPW.}

We evaluate DensePose~3D in a transfer-learning setting, training it on Human~3.6M and evaluating it on 3DPW~\cite{marcard18recovering}.
DP3D takes keypoints as input, so is invariant to appearance changes and generalises well, as can be seen in~\Cref{t:quant,f:h36mres}.
We follow the same evaluation protocol as for Human~3.6M, comparing RE on 14 joints.

\paragraph{LVIS.}

Finally, we fit our model to LVIS dataset~\cite{gupta2019lvis} containing animal images taken ``in the wild''.
This task is more challenging, since each category comprises only about 2000 training instances, many of which have occluded parts.
To get input keypoints and visibilities~$(\Y,\Z)$, we pre-process the images with CSE~\cite{neverova2020continuous} in a similar way to DensePose.
%
%
The output of CSE is noisier than the one of DensePose, so we predict heteroscedastic variance for the loss~\eqref{e:rep} and maximise the log-likelihood of the Laplace distribution as done by Novotny~\etal~\cite{novotny18capturing}; see sup.~mat.~for details on pre-processing and the loss.
Since there is no 3D ground truth, we provide only qualitative results in~\Cref{f:lvisres}.

\subsection{Comparison to baselines}

\begin{table}[tb]
\centering
\begin{tabular}{l|rrrr}
\toprule
Method                                    & UP-3D         & H3.6M          & 3DPW  & Dogs   \\ \midrule
HMR \cite{kanazawa18end-to-end}           & ---           & 56.8           & 81.3  & ---    \\
GraphCMR \cite{kolotouros19convolutional} & ---           & 50.1           & 70.2  & ---    \\
SPIN \cite{kolotouros19learning}          & ---           & 41.8           & 59.3  & ---    \\
Multi-bodies \cite{biggs20203d}           & ---           & 46.1           & 59.9  & ---    \\ \midrule
C3DPO~\cite{novotny19c3dpo}                  & 107.0         & 216.6          & 199.9 & 345.1    \\
no canon.\,loss~\eqref{e:canon}           & 183.6          & 135.4 & 120.3 & \textbf{241.4}     \\
no ARAP loss~\eqref{e:arap}               & 242.6          & 154.8 & 126.1 & 371.8  \\
no entropy loss~\eqref{e:entropy}         & 113.8          & 119.4 & 99.1  & 505.2  \\
no parts model                            & 205.9         & 125.0          & 102.3 & 684.3   \\
\method (ours)                            & \textbf{91.2} & \textbf{113.6} & \textbf{95.2} &  247.1   \\ \bottomrule
\end{tabular}
\caption{%
\textbf{Evaluation of mesh reconstruction}
reporting mean per-joint position error (MPJPE) on UP-3D and Dogs datasets, and reconstruction error (RE) on Human 3.6M and 3DPW.
The first half of the table shows the results of methods that use 3D supervision.
\method is then compared to C3DPO~\cite{novotny19c3dpo} applied to dense keypoints
and ablated.
}\label{t:quant}
\end{table}

We compare our method to C3DPO~\cite{novotny19c3dpo}, where we use 10-dimensional basis and find the optimal strength of canonicalisation loss in the interval~$[0.1, 1]$.
The results are in~\Cref{t:quant} and supplementary figures.
Note that we train C3DPO on dense keypoints (i.e.~6890 input points for humans), while \cite{novotny19c3dpo} trains on 17 sparse joints, which makes results from \Cref{t:quant} incomparable with the ones in \cite{novotny19c3dpo}.
UP-3D and Dogs are less-challenging datasets with clean 2D keypoints and few extreme poses, so C3DPO's simple linear pose model is only slightly inferior to DP3D.
In contrast, the gap is large on Human~3.6M and 3DPW: C3DPO outputs the mean pose failing to adapt to the data.

\graphicspath{{images/qualitative/}}
\newcommand{\cdpodir}{dpc3dpo-h36m21_017_c3dpo_00000_MOD.camera_scale=True_M.l.l_canonicalization=0.1}
\newcommand{\oursdir}{dpc3dpo-h36m21_021_weight_repro_00003_M.l.l_arap=0.3_MOD.shared_scale=False}
\newcommand{\lbocdpodir}{dpc3dpo-h36m21_018_lboc3dpo_00000_MOD.lbo_vertex_transform_model=translation_M.l.l_canonicalization=1.0_MOD.shape_basis_size=10}

\newcommand{\imh}{1.72cm}
\newcommand{\linegap}{\vspace{-0.0cm}}

\newcommand{\methodfig}[2]{%
\includegraphics[trim=2cm 0.5cm 2cm 3.2cm, clip, height=\imh]{#1/#2.png}%
}

\newcommand{\cdpofig}[2]{%
\includegraphics[trim=1cm 0cm 1cm 1.7cm, clip, height=\imh]{#1/#2.png}%
}

\newcommand{\imfig}[2]{%
\includegraphics[height=\imh]{#1/#2_image_kp_partseg.png}%
}

\newcommand{\rainbowours}{%
{\color[HTML]{8D274B}D}%
{\color[HTML]{A48053}P}%
{\color[HTML]{547CA2}3}%
{\color[HTML]{8EA395}D}%
}

\newcommand{\methtext}[1]{\centering\scriptsize\hspace{0.01cm}\rotatebox{90}{\hspace{0.1cm}#1}\hspace{0.01cm}}

\newcommand{\comparerow}[2]{%
\begin{minipage}[c]{0.24\linewidth}\centering%
    \imfig{#2\oursdir}{#1}\\\linegap
    \methtext{~~~~w/o parts}%
    \methodfig{#2\lbocdpodir}{#1_mesh_over_image_raw_solid}%
    \methodfig{#2\lbocdpodir}{#1_mesh_raw_solid_000_089}\\\linegap
    \methtext{~\textbf{\rainbowours~\color[HTML]{000000}(ours)}}%
    \methodfig{#2\oursdir}{#1_mesh_over_image_raw_partseg}%
    \methodfig{#2\oursdir}{#1_mesh_raw_partseg_000_089}%
\end{minipage}}

\begin{figure*}
\comparerow{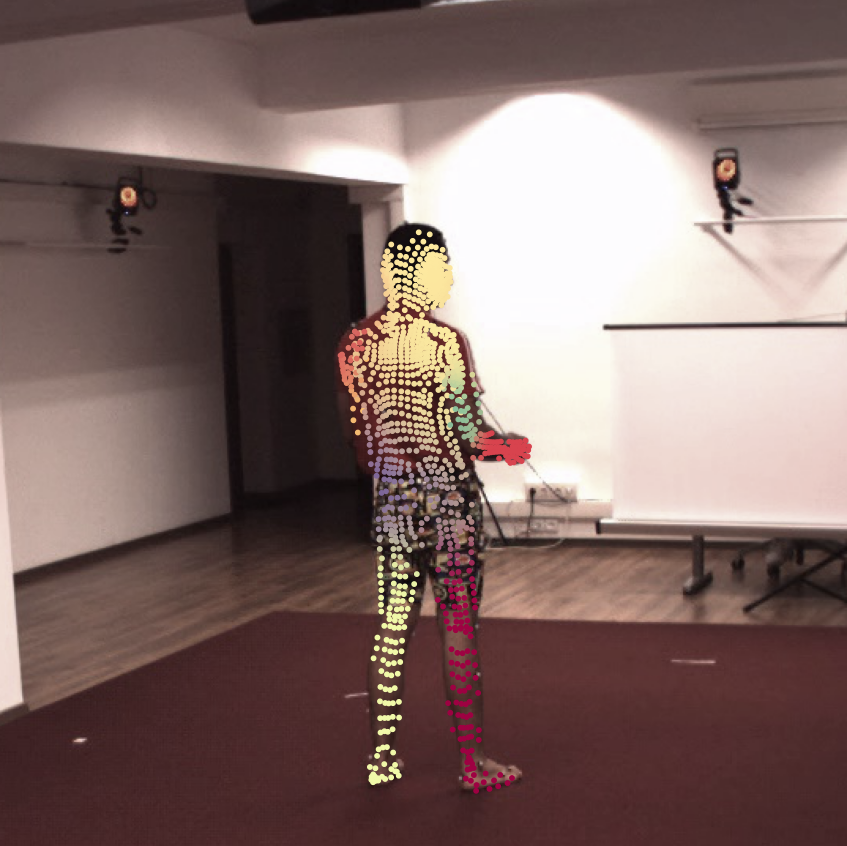}{}
\comparerow{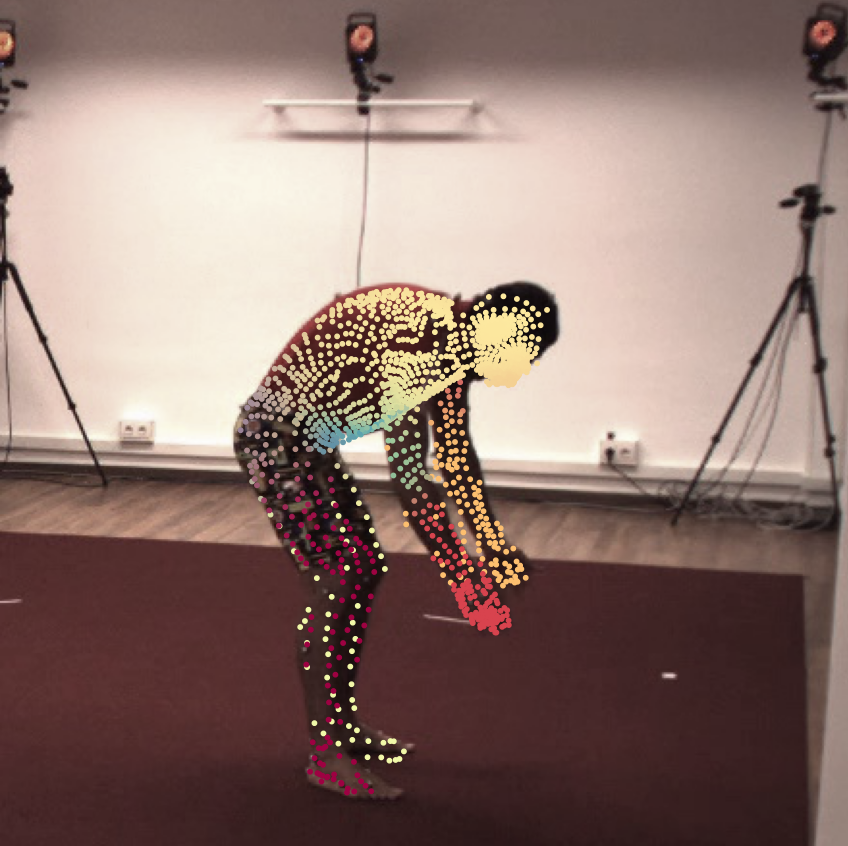}{}
\comparerow{downtown_runForBus_00/image_00090}{}
\comparerow{downtown_weeklyMarket_00/image_00532}{}

\caption{\textbf{Qualitative evaluation on Human 3.6M (left two images) and 3DPW (right two images)}.
From top to bottom: input image and keypoints, reconstruction with the linear model instead of parts segmentation,
and of the proposed method.%
}\label{f:h36mres}
\end{figure*}

\let\cdpodir\undefined
\let\lbocdpodir\undefined
\let\oursdir\undefined
\let\imh\undefined
\let\cdpodir\undefined
\let\oursdir\undefined
\let\imh\undefined
\let\methodfig\undefined
\let\imfig\undefined
\let\rainbowours\undefined
\let\methtext\undefined
\let\comparerow\undefined
\let\linegap\undefined
\let\cdpofig\undefined
\let\imfig\undefined
\let\rainbowours\undefined

\subsection{Ablation study}

\paragraph{Removing the loss functions.}

We report the effect of removing various regularisers in \Cref{t:quant}.
Each of them proves important:
the canonicalisation loss prevents predicting a degenerate, flat shape;
the ARAP loss makes the prediction smooth and helps to learn smooth part segmentations by encouraging local rigidity;
the entropy loss makes the part segmentation sharper, allowing the shape to flex more.

\graphicspath{{images/qualitative/}}
\newcommand{\oursdir}{dpc3dpo-lvis_029_sigma_zebra_00024_M.l.l_local_rigidity=0.001_M.l.l_arap=1.0_MOD.num_latent_parts=10_MOD.uncertainty_sigma_clip=0.1}
\newcommand{\lbocdpodir}{dpc3dpo-lvis_043_zebra_5002_lboc3dpo_00000_M.l.l_canonicalization=1.0_MOD.shape_basis_size=10}

\newcommand{\oursdirbear}{dpc3dpo-lvis_024_bear_4936_00012_M.l.l_arap=1.0_MOD.num_latent_parts=7_MOD.uncertainty_sigma_clip=0.1}
\newcommand{\lbocdpodirbear}{dpc3dpo-lvis_037_bear_4936_lboc3dpo_00000_M.l.l_canonicalization=1.0_MOD.shape_basis_size=10}

\newcommand{\imh}{1.72cm}
\newcommand{\linegap}{\vspace{-0.0cm}}

\newcommand{\methodfigbg}[3]{%
\includegraphics[trim=0.0cm 0.0cm 0.0cm 0.0cm, clip, height=\imh]{#1/#2_mesh_over_image_raw_#3.png}%
}
\newcommand{\methodfig}[3]{%
\includegraphics[trim=1cm 0.25cm 1cm 1.6cm, clip, height=\imh]{#1/#2_mesh_raw_#3_000_089.png}%
}

\newcommand{\lbocdpofigbg}[1]{%
\includegraphics[trim=0.0cm 0.0cm 0.0cm 0.0cm, clip, height=\imh]{\lbocdpodir/#1_mesh_over_image_raw_solid.png}%
}
\newcommand{\lbocdpofig}[1]{%
\includegraphics[trim=1cm 0.25cm 1cm 1.6cm, clip, height=\imh]{\lbocdpodir/#1_mesh_raw_solid_000_089.png}%
}

\newcommand{\imfig}[1]{%
\includegraphics[height=\imh]{\oursdir/#1_image_kp_partseg.png}%
}

\newcommand{\imfigbear}[1]{%
\includegraphics[height=\imh]{\oursdirbear/#1_image_kp_partseg.png}%
}

\newcommand{\rainbowours}{%
{\color[HTML]{8D274B}D}%
{\color[HTML]{A48053}P}%
{\color[HTML]{547CA2}3}%
{\color[HTML]{8EA395}D}%
}

\newcommand{\methtext}[1]{\centering\scriptsize\hspace{0.01cm}\rotatebox{90}{\hspace{0.1cm}#1}\hspace{0.01cm}}

\newcommand{\comparerow}[1]{%
\begin{minipage}[c]{0.24\linewidth}\centering%
    \imfig{#1}\\\linegap
    \methtext{~~~~w/o parts}%
    \methodfigbg{\lbocdpodir}{#1}{solid}%
    \methodfig{\lbocdpodir}{#1}{solid}\\\linegap%
    \methtext{~\textbf{\rainbowours~\color[HTML]{000000}(ours)}}%
    \methodfigbg{\oursdir}{#1}{partseg}%
    \methodfig{\oursdir}{#1}{partseg}%
\end{minipage}}

\newcommand{\comparerowbear}[1]{%
\begin{minipage}[c]{0.24\linewidth}\centering%
    \imfigbear{#1}\\\linegap
    \methtext{~~~~w/o parts}%
    \methodfigbg{\lbocdpodirbear}{#1}{solid}%
    \methodfig{\lbocdpodirbear}{#1}{solid}\\\linegap%
    \methtext{~\textbf{\rainbowours~\color[HTML]{000000}(ours)}}%
    \methodfigbg{\oursdirbear}{#1}{partseg}%
    \methodfig{\oursdirbear}{#1}{partseg}%
\end{minipage}}

\begin{figure*}
\comparerowbear{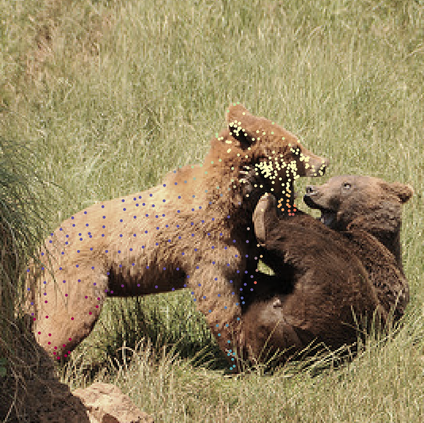}
\comparerowbear{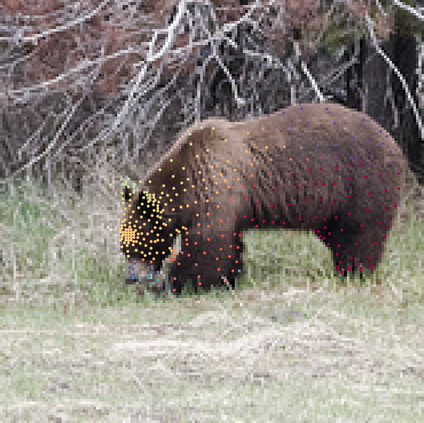}
\comparerow{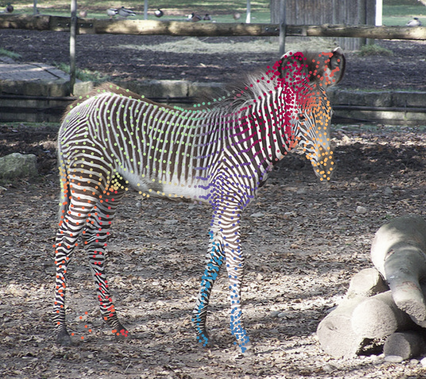}
\comparerow{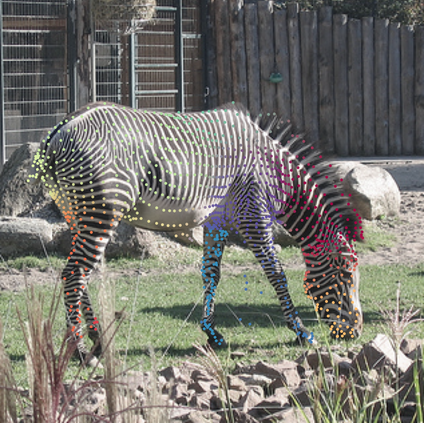}

\caption{\textbf{Qualitative evaluation on LVIS}.
From top to bottom: input image and keypoints,
the reconstruction with the linear model instead of parts segmentation,
and of the proposed method, where colours correspond to the learnt part segmentation.%
\vspace{-0.5cm}%
}\label{f:lvisres}
\end{figure*}

\let\cdpodir\undefined
\let\lbocdpodir\undefined
\let\oursdir\undefined
\let\methodfigbg\undefined
\let\lbocdpofigbg\undefined
\let\lbocdpodirbear\undefined
\let\oursdirbear\undefined
\let\imh\undefined
\let\cdpodir\undefined
\let\oursdir\undefined
\let\imh\undefined
\let\methodfig\undefined
\let\lbocdpofig\undefined
\let\imfig\undefined
\let\rainbowours\undefined
\let\methtext\undefined
\let\comparerow\undefined
\let\comparerowbear\undefined
\let\linegap\undefined
\let\cdpofig\undefined
\let\imfigbear\undefined
\let\rainbowours\undefined

\paragraph{Removing the part-based model.}

Two reasons why C3DPO may work poorly on dense point clouds are:
(1) learning a very large linear predictor for thousand of points may lead to overfitting, or
(2) the linear model may be unable to capture surface deformations.
We test these hypotheses by replacing the articulation model in our method with a C3DPO-like linear basis.
To reduce the number of parameters in the basis, we express it as a function of the LBO basis $\mathbf{U}$ (\cref{s:method}) and define the posed mesh as
$\label{e:lboc3dpo}
\X = (\alpha \otimes I_3) \mathbf{W}^b \mathbf{U},
$
where $\mathbf{W}^b \in \mathbb{R}^{D \times N_u}$ are trainable parameters, $\alpha$ is a $D$-dimensional vecror of shape coefficients, $\otimes$ is Kronecker product, and $I_3$ is a 3-dimensional identity matrix.
We train using~\cref{e:totalloss} but remove the entropy loss~\eqref{e:entropy} (as this model has no parts).
We set the number of blendshapes $D=10$ and find the optimal weight of canonicalisation loss in the $[0.1, 1]$ range.

The penultimate row in~\Cref{t:quant} reveals the correct hypothesis.
The model without parts performs significantly better than C3DPO, proving that overfitting explains in large part C3DPO's poor performance.
However, the no-parts model still cannot reach the performance of \method on real-world data (columns 2 and 3), which means that the latter is more efficient than using vanilla linear blendshapes.
Remarkably, the no-parts model performs decently on the synthetic datasets, where  DensePose  annotations  were  simulated  by projecting 3D locations obtained from a parametric model, probably because, despite of high dimensionality, the ``rank'' of the data is still small.
The differences are more pronounced in the visual results in \Cref{f:h36mres,f:lvisres,f:dogsres}.
The linear model in most cases produces symmetrical shapes, which tend to be similar regardless of the input, while DP3D with parts reconstructs the movements of arms more accurately.\vspace{-0.5ex}

\begin{figure}[b!]
    \includegraphics[width=0.99\linewidth]{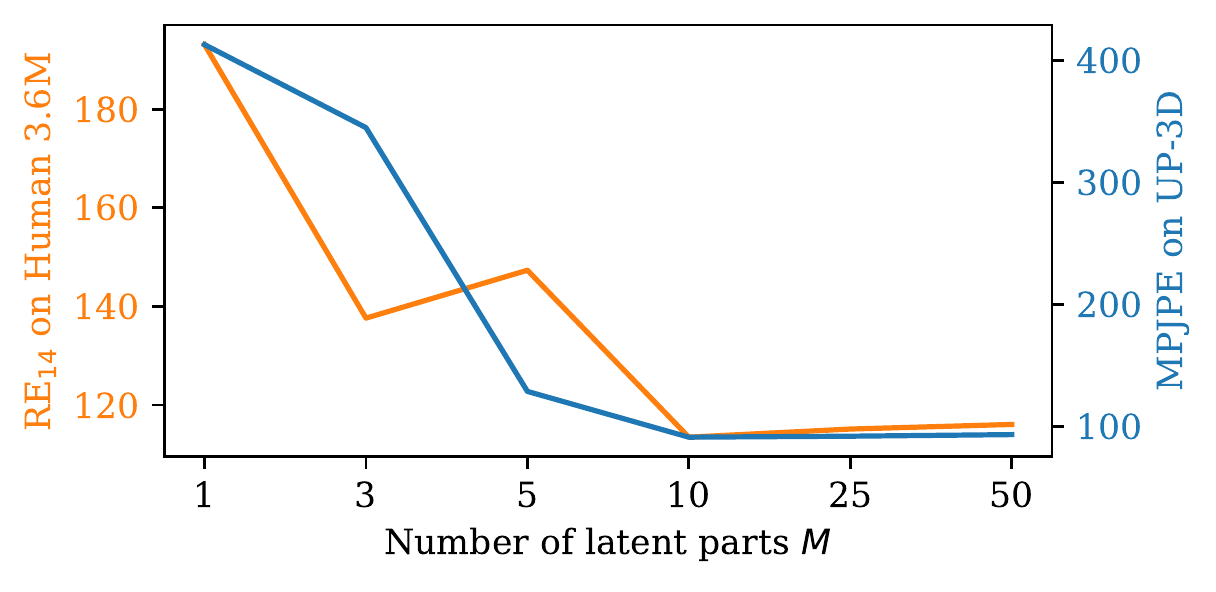}
    \caption{%
    Reconstruction quality w.r.t.~the number of parts.%
    }
    \label{f:varnumparts}
\end{figure}

\paragraph{Number of latent parts.}

\Cref{f:varnumparts} measures the reconstruction error as a function of the number of latent parts~$M$ on human datasets.
As expected from human anatomy, the method needs at least 5 parts to model the articulation of the body.
The metrics plateau after 10 parts.\vspace{-0.5ex}

\paragraph{Limitations and robustness.}
DensePose\,3D can be only as good as training annotations provided by DensePose or CSE.
In supp.~mat., we investigate the sensitivity of training to annotation noise, random sparsity (typical for manual annotation), and missing body parts (caused by occlusions).

\section{Conclusions}\label{s:conclusion}

We presented a method that learns 3D deformable shape reconstruction given only a single artist-generated rigid template mesh and dense 2D keypoint annotations,
without the need for 3D supervision with the deformable shape model or 3D pose regressor, which are difficult to obtain for most object categories.
Because of this, we apply DP3D to the reconstruct animals that lack such 3D annotations.%
\ifdefined\arxivsubmission%
\clearpage
\appendix
\setcounter{figure}{0} \renewcommand{\thefigure}{\Roman{figure}}
\setcounter{table}{0} \renewcommand{\thetable}{\Roman{table}}
\begin{strip}%
 \centering
 \Large
 \textbf{%
 DensePose 3D\@: Lifting Canonical Surface Maps of Articulated Objects to the Third Dimension\\
 \vspace{0.3cm} \textit{Supplementary material}
 }\\
\end{strip}
\section{Implementation details}

We implemented our model using Pytorch (the code will be shared).
The networks~$\Phi$ in~eq.~(4) and~$\Psi$~(6) have the same architecture as in C3DPO~\cite{Novotny2019}.
In particular, they first map the input to a 1024-dimensional vector with a fully-connected layer followed by 6 residual 3-layer MLPs.
The blocks have the architecture 1024--256--256--1024, each layer followed by BatchNorm and ReLU.
The network parameters are optimised with SGD with momentum and weight decay.
Training runs for 20 epochs on Human~3.6M, for 100 epochs on UP-3D, 200 epochs for Stanford Dogs, and 1000 epochs for the small LVIS categories, starting with the learning rate 0.003 and decreasing 10 times after 80\% of epochs have passed.
The momentum coefficient is 0.99 and weight decay to 0.001.
The network takes sparse keypoints and is thus lean on memory compared to convolutional neural networks, which have to maintain feature maps,
thus a batch size of 512 can be used with a 16 GB GPU.
A forward pass for this batch size takes 0.5 seconds.

To project the predicted 3D shape back to the image plane in loss~(5), we define~$\pi$ as orthographic projection,
although the method allows using perspective projection as long as camera intrinsics are known.
The input keypoints~$\Y$ are zero-centered before passing to~$\Phi$.

The hyperparameters were set to the following values:
the number of latent parts~$M$ in the segmentation model was set to 10;
segmentation model initialisation parameter~$\bar{\sigma} = 32$,
weights of the loss functions in~eq.~(9) are:
$w_{entropy} = 0.001$,
$w_{arap} = 0.3$,
$w_{canon} = 0.1$.

\section{Generating keypoints for LVIS}

For reconstruction of humans, we pre-process data with DensePose~\cite{guler18densepose:} to obtain UV coordinates defining the correspondences to the template mesh, 
and convert them to 2D keypoints corresponding to template mesh vertices as described in Section 3.1.
For animals we instead use pre-trained CSE models~\cite{neverova20continuous} to obtain the per-pixel descriptors from the joint embedding space with category-specific template mesh surface.
For each pixel within the object mask, we find the closest template mesh vertex in the embedding space.
Let the set of pixels~$j$ that were matched to the vertex~$i$ of the template be
\begin{equation}
\mathcal{M}_i = \{j : i = \textrm{argmin}_{i'}~  \|\mathbf{e}_j - \mathbf{e}_{i'}\| \},
\end{equation}
where $\mathbf{e}_j$ and $\mathbf{e}_{i'}$ are the embeddings of the $j$-th pixel and $i'$-th vertex, respectively.
Then, all vertices that have been matched to at least one pixel, i.e.~$\mathcal{M}_i \ne \emptyset$, are considered visible.
For each visible vertex, we find the corresponding 2D keypoint location as the mean coordinate of the pixels matched to it: $y_i = \mathsf{E}_{j \in \mathcal{M}_i}~ p_j$, where~$p_j$ are the coordinates in the pixel grid.
This way, we ensure that all occluded surface points are marked as invisible in the DensePose~3D input.

\section{Heteroscedastic reprojection loss}

The annotation by CSE~\cite{neverova20continuous} trained on LVIS is quite noisy due to small dataset size and high amount of noise and occlusions.
When fitting the model, it is generally important to use a robust loss such as L1 or Huber (we use pseudo-Huber loss in this paper wherever the norm is not explicitly specified).
For CSE annotation specifically, we found important to predict the keypoint-specific variance to weigh its contribution to the loss.

To properly account for variance, we generalise L1 loss in correspondence with the maximum likelihood estimation theory similar to what Novotny~\etal~\cite{novotny18capturing} did for L2 loss.
First, note that L1 loss is proportional to a shifted negative log-likelihood of Laplacian distribution of the residual's L1 norm, given constant scale parameter~$b$.
In the heteroscedastic version, we do not fix those additive and multiplicative terms and consider the full NLL:

\begin{equation}\label{e:rep}
-\log p(y | \hat{y}, b) =
\log(2b) + \frac{|y - \hat{y}|}{b}.
\end{equation}

For numerical stability, the denominator is clipped, and the constant term is removed, which brings us to the following formulation:
\begin{equation}\label{e:rep}
\mathcal{L}^b_\text{rep} (\hat{y}, y, b) = 
\log b + \frac{\mathcal{L}_\text{rep}(\hat{y}, y)}{\max\{b, b_{\textrm{min}}\}},
\end{equation}
where~$b_{\textrm{min}}$ is set to 0.1, and~$\mathcal{L}_\text{rep}(\hat{y}, y)$ is a pseudo-Huber loss $\epsilon (\sqrt{1 + (\|y - \hat{y}\|/\epsilon)^2} - 1)$, which smoothly approximates the L1 loss.

Finally, the per-keypoint uncertainty $b_k$~is also predicted by the model based on keypoint's identity and its predicted local pose.
We train a new branch that takes for each keypoint~$k$ its LBO descriptor concatenated with the predicted transformation of the corresponding template vertex. It comprises a 2-layer MLP topped with softplus.
The LBO descriptor identifies the keypoint in a smooth way, while the transformation can aid uncerainty prediction because the 2D projection of the variance depends on the angle between the surface element and the camera ray.

\section{Data and model for dogs}
Human shape models like SMPL combine linear blendshapes with linear blend skinning; the former is responsible for modelling the body type, while the letter for articulation.
While we can do the same in our model, we found that for humans it did not yield any improvement.
The variation in the shape of dogs across breeds is in contrast huge, e.g.~the body lenght to leg lenght ratio is different for great danes versus dachshunds.
To model this shape variations, we learn a set of 5 blendshapes and apply them to the template mesh before posing it.
The blendshapes are learned as a vertex-wise function of the LBO basis, in the same way as in our no-parts baseline described in Section~4.3.
This parametrisation of blendshapes as a linear function of LBO descriptors again helps to achieve invariance to remeshing and avoid overfitting.

When generating the dataset, we remove cases where less than 10\% of the template surface area is visible because this usually results in poor SMAL fits.

\section{Robustness of training}

\begin{figure}[tb]
  \includegraphics[width=0.9\linewidth]{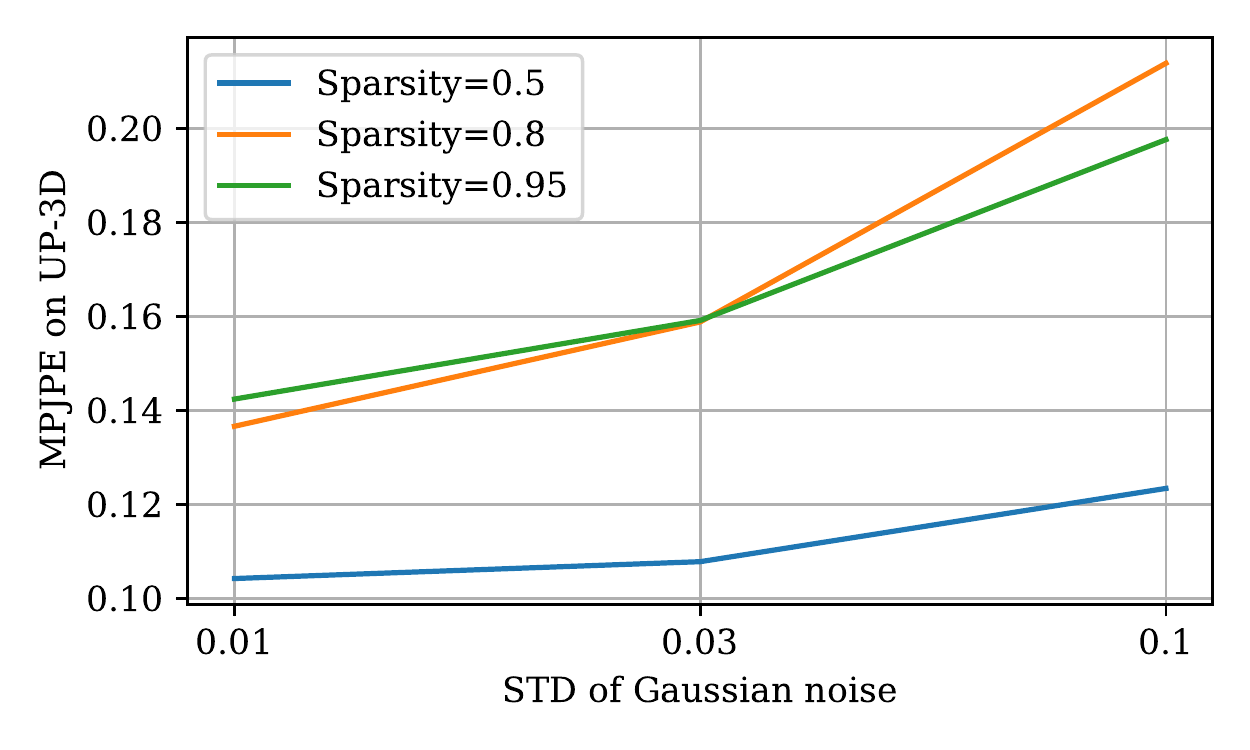}\vspace{-3ex}%
  \caption{%
  Reconstruction quality w.r.t.~the noise for different sparsity.%
  }
  \label{f:dp3dnoise}
\end{figure}

\begin{figure}[tb]
  \includegraphics[width=0.9\linewidth]{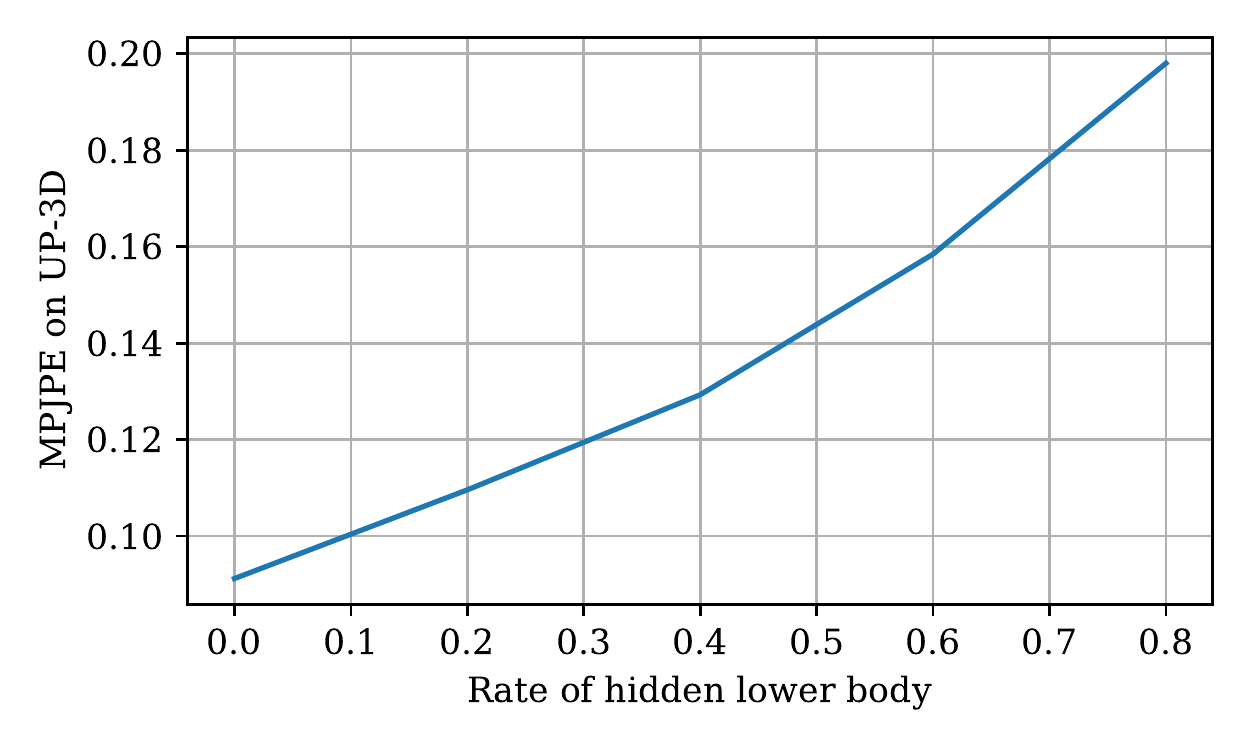}\vspace{-3ex}%
  \caption{%
  Reconstruction quality w.r.t.~the rate of removed legs.%
  }
  \label{f:dp3d-legs}
\end{figure}

\method training relies on roughly correct DensePose predictions. In case DensePose failed (e.g. due to occlusions), our model would not be able to recover.
Here we investigate how the training is robust to different kinds of noise added to synthetic UP-3D. First, we simulate the case of manual annotations by adding Gaussian noise to 2D keypoints and marking some portions of them as invisible. \Cref{f:dp3dnoise} shows that adding uncorrelated Gaussian noise is not harmful as long as keypoints are relatively dense, but it becomes a problem once the location of 80\%+ of projected keypoints is unknown. Second, in \Cref{f:dp3d-legs}, we marked invisible the whole lower half of the body (in the canonical orientation) for a certain proportion of training instances. As expected, the method is less robust to occlusions than to Gaussian noise.

\section{More qualitative results}

See more results in comparison to baselines in~\Cref{f:lvisres-sup,f:h36mres-sup,f:3dpwres-sup,f:dogsres-sup}.

\graphicspath{{images/qualitative/}}
\newcommand{\oursdir}{dpc3dpo-lvis_029_sigma_zebra_00024_M.l.l_local_rigidity=0.001_M.l.l_arap=1.0_MOD.num_latent_parts=10_MOD.uncertainty_sigma_clip=0.1}
\newcommand{\lbocdpodir}{dpc3dpo-lvis_043_zebra_5002_lboc3dpo_00000_M.l.l_canonicalization=1.0_MOD.shape_basis_size=10}

\newcommand{\oursdirbear}{dpc3dpo-lvis_024_bear_4936_00012_M.l.l_arap=1.0_MOD.num_latent_parts=7_MOD.uncertainty_sigma_clip=0.1}
\newcommand{\lbocdpodirbear}{dpc3dpo-lvis_037_bear_4936_lboc3dpo_00000_M.l.l_canonicalization=1.0_MOD.shape_basis_size=10}

\newcommand{\imh}{1.72cm}
\newcommand{\linegap}{\vspace{-0.0cm}}

\newcommand{\methodfigbg}[3]{%
\includegraphics[trim=0.0cm 0.0cm 0.0cm 0.0cm, clip, height=\imh]{#1/#2_mesh_over_image_raw_#3.png}%
}
\newcommand{\methodfig}[3]{%
\includegraphics[trim=1cm 0.25cm 1cm 1.6cm, clip, height=\imh]{#1/#2_mesh_raw_#3_000_089.png}%
}

\newcommand{\lbocdpofigbg}[1]{%
\includegraphics[trim=0.0cm 0.0cm 0.0cm 0.0cm, clip, height=\imh]{\lbocdpodir/#1_mesh_over_image_raw_solid.png}%
}
\newcommand{\lbocdpofig}[1]{%
\includegraphics[1cm 0.25cm 1cm 1.6cm, clip, height=\imh]{\lbocdpodir/#1_mesh_raw_solid_000_089.png}%
}

\newcommand{\imfig}[1]{%
\includegraphics[height=\imh]{\oursdir/#1_image_kp_partseg.png}%
}

\newcommand{\imfigbear}[1]{%
\includegraphics[height=\imh]{\oursdirbear/#1_image_kp_partseg.png}%
}

\newcommand{\rainbowours}{%
{\color[HTML]{8D274B}D}%
{\color[HTML]{A48053}P}%
{\color[HTML]{547CA2}3}%
{\color[HTML]{8EA395}D}%
}

\newcommand{\methtext}[1]{\centering\scriptsize\hspace{0.01cm}\rotatebox{90}{\hspace{0.1cm}#1}\hspace{0.01cm}}

\newcommand{\comparerow}[1]{%
\begin{minipage}[c]{0.24\linewidth}\centering%
    \imfig{#1}\\\linegap
    \methtext{~~~~w/o parts}%
    \methodfigbg{\lbocdpodir}{#1}{solid}%
    \methodfig{\lbocdpodir}{#1}{solid}\\\linegap%
    \methtext{~\textbf{\rainbowours~\color[HTML]{000000}(ours)}}%
    \methodfigbg{\oursdir}{#1}{partseg}%
    \methodfig{\oursdir}{#1}{partseg}%
\end{minipage}}

\newcommand{\comparerowbear}[1]{%
\begin{minipage}[c]{0.24\linewidth}\centering%
    \imfigbear{#1}\\\linegap
    \methtext{~~~~w/o parts}%
    \methodfigbg{\lbocdpodirbear}{#1}{solid}%
    \methodfig{\lbocdpodirbear}{#1}{solid}\\\linegap%
    \methtext{~\textbf{\rainbowours~\color[HTML]{000000}(ours)}}%
    \methodfigbg{\oursdirbear}{#1}{partseg}%
    \methodfig{\oursdirbear}{#1}{partseg}%
\end{minipage}}

\begin{figure*}
\comparerowbear{000000377282}
\comparerowbear{000000218960}
\comparerowbear{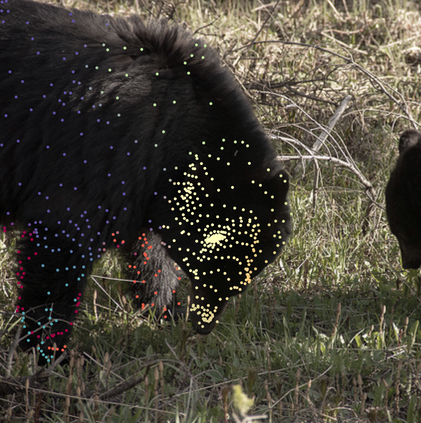}
\comparerowbear{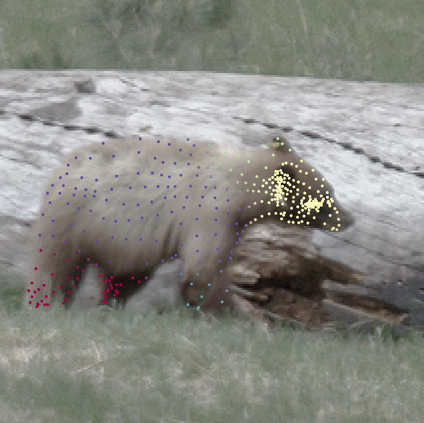}\\\linegap
\comparerow{000000229107}
\comparerow{000000187434}
\comparerow{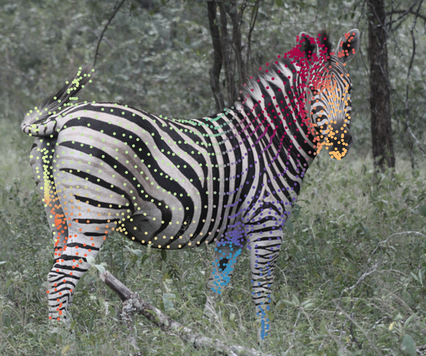}
\comparerow{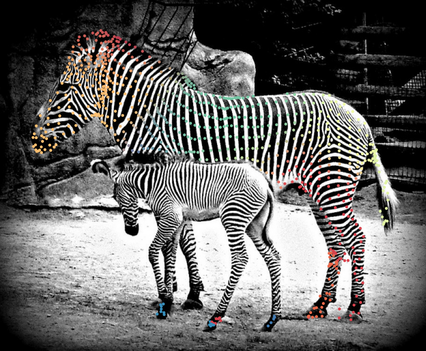}

\caption{\textbf{Qualitative evaluation on LVIS}.
Each of the rows contains, from top to bottom: input image and keypoints,
the reconstruction with the linear model instead of parts segmentation,
and of the proposed method.%
}\label{f:lvisres-sup}
\end{figure*}

\let\cdpodir\undefined
\let\lbocdpodir\undefined
\let\oursdir\undefined
\let\imh\undefined
\let\noentrdir\undefined
\let\noarapdir\undefined
\let\nocandir\undefined
\let\methodfig\undefined
\let\imfig\undefined
\let\rainbowours\undefined
\let\methtext\undefined
\let\comparerow\undefined
\let\linegap\undefined
\let\cdpofig\undefined
\let\imfig\undefined
\let\rainbowours\undefined

\graphicspath{{images/qualitative/}}
\newcommand{\cdpodir}{dpc3dpo-h36m21_017_c3dpo_00000_MOD.camera_scale=True_M.l.l_canonicalization=0.1}
\newcommand{\oursdir}{dpc3dpo-h36m21_021_weight_repro_00003_M.l.l_arap=0.3_MOD.shared_scale=False}
\newcommand{\lbocdpodir}{dpc3dpo-h36m21_018_lboc3dpo_00000_MOD.lbo_vertex_transform_model=translation_M.l.l_canonicalization=1.0_MOD.shape_basis_size=10}
\newcommand{\noarapdir}{dpc3dpo-h36m21_025_reprow_ablation_00003_M.l.l_canonicalization=0.1_M.l.l_part_entropy=0.001_M.l.l_arap=0.0_MOD.shared_scale=False}
\newcommand{\noentrdir}{dpc3dpo-h36m21_025_reprow_ablation_00005_M.l.l_canonicalization=0.1_M.l.l_part_entropy=0.0_M.l.l_arap=0.3_MOD.shared_scale=False}
\newcommand{\nocandir}{dpc3dpo-h36m21_025_reprow_ablation_00009_M.l.l_canonicalization=0.0_M.l.l_part_entropy=0.001_M.l.l_arap=0.3_MOD.shared_scale=False}

\newcommand{\imh}{1.72cm}
\newcommand{\linegap}{\vspace{-0.0cm}}

\newcommand{\methodfig}[2]{%
\includegraphics[trim=2cm 0.5cm 2cm 3.2cm, clip, height=\imh]{#1/#2.png}%
}

\newcommand{\cdpofig}[2]{%
\includegraphics[trim=1cm 0cm 1cm 1.7cm, clip, height=\imh]{#1/#2.png}%
}

\newcommand{\imfig}[1]{%
\includegraphics[height=\imh]{\oursdir/#1_image_kp_partseg.png}%
}

\newcommand{\rainbowours}{%
{\color[HTML]{8D274B}D}%
{\color[HTML]{A48053}P}%
{\color[HTML]{547CA2}3}%
{\color[HTML]{8EA395}D}%
}

\newcommand{\methtext}[1]{\centering\scriptsize\hspace{0.01cm}\rotatebox{90}{\hspace{0.1cm}#1}\hspace{0.01cm}}

\newcommand{\comparerow}[1]{%
\begin{minipage}[c]{0.24\linewidth}\centering%
    \imfig{#1}\\\linegap
    \methtext{~\textbf{\color[HTML]{960061}C3DPO \cite{Novotny2019}}}%
    \cdpofig{\cdpodir}{#1_pcl_lbo_filtered_solid_000_000}%
    \cdpofig{\cdpodir}{#1_pcl_lbo_filtered_solid_000_089}\\\linegap
    \methtext{~~~w/o ARAP}%
    \methodfig{\noarapdir}{#1_mesh_raw_partseg_000_000}%
    \methodfig{\noarapdir}{#1_mesh_raw_partseg_000_089}\\\linegap
    \methtext{w/o canonical'n}%
    \methodfig{\nocandir}{#1_mesh_raw_partseg_000_000}%
    \methodfig{\nocandir}{#1_mesh_raw_partseg_000_089}\\\linegap
    \methtext{~~w/o entropy}%
    \methodfig{\noentrdir}{#1_mesh_raw_partseg_000_000}%
    \methodfig{\noentrdir}{#1_mesh_raw_partseg_000_089}\\\linegap
    \methtext{~~~~w/o parts}%
    \methodfig{\lbocdpodir}{#1_mesh_raw_solid_000_000}%
    \methodfig{\lbocdpodir}{#1_mesh_raw_solid_000_089}\\\linegap
    \methtext{~\textbf{\rainbowours~\color[HTML]{000000}(ours)}}%
    \methodfig{\oursdir}{#1_mesh_raw_partseg_000_000}%
    \methodfig{\oursdir}{#1_mesh_raw_partseg_000_089}%
\end{minipage}}


\begin{figure*}
\comparerow{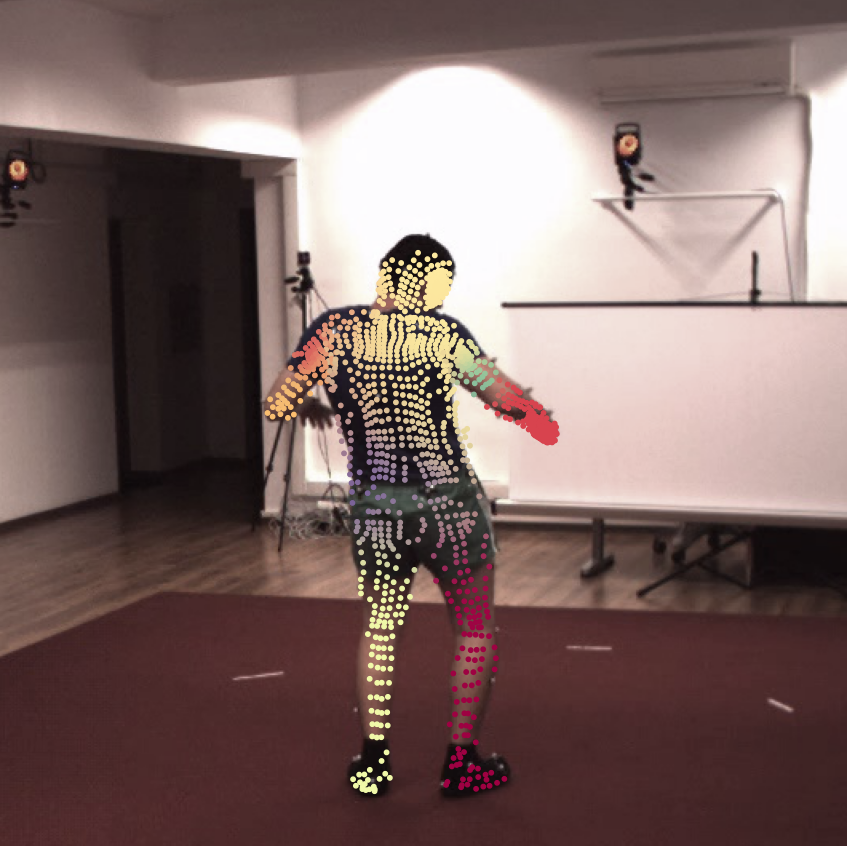}
\comparerow{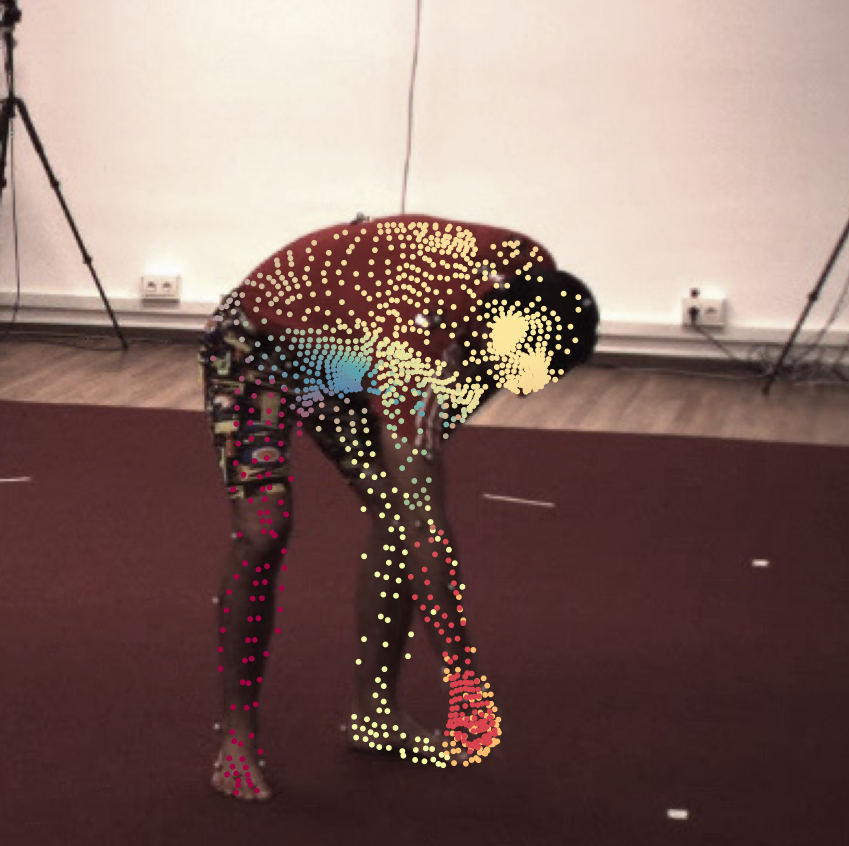}
\comparerow{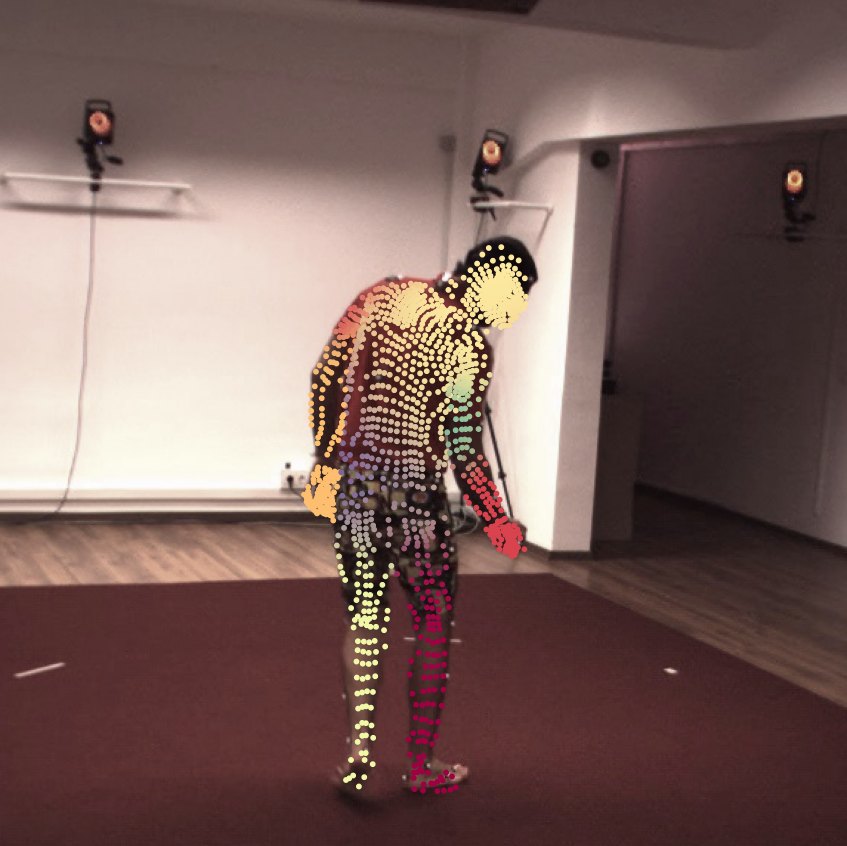}
\comparerow{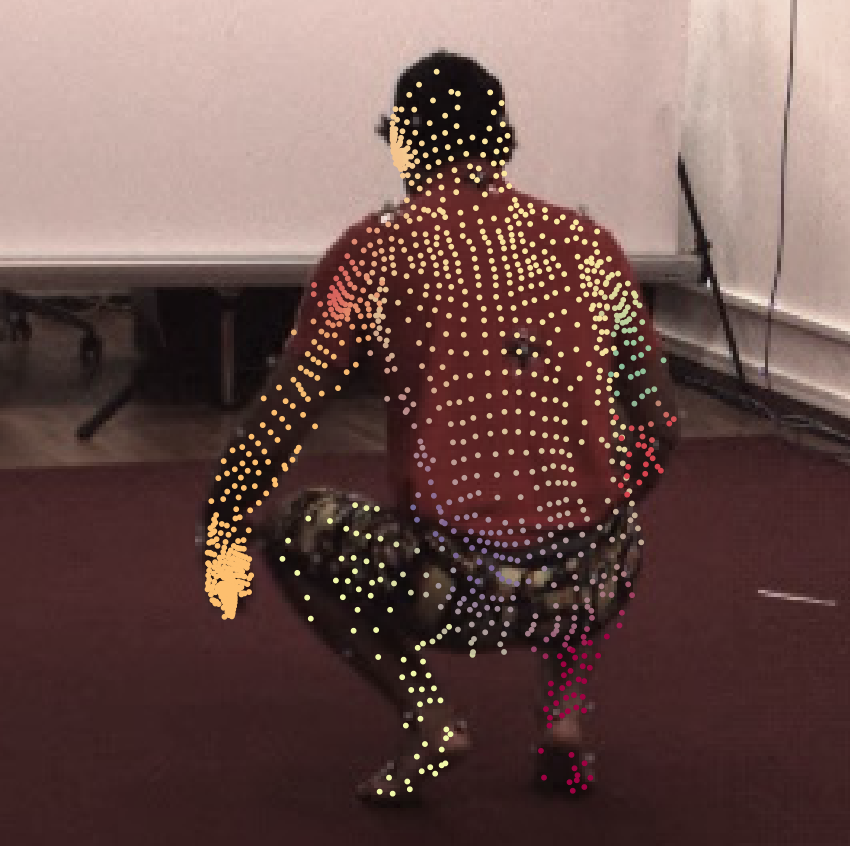}\\

\caption{\textbf{Qualitative evaluation on Human 3.6M}.
From top to bottom: input image and keypoints,
the reconstruction of C3DPO~\cite{Novotny2019}, of DensePose~3D without the ARAP loss~(7),
without the canonicalisation loss~(6), without the entropy loss~(8), with the linear model instead of parts segmentation,
and of the full proposed method.%
}\label{f:h36mres-sup}
\end{figure*}

\let\cdpodir\undefined
\let\lbocdpodir\undefined
\let\oursdir\undefined
\let\imh\undefined
\let\noentrdir\undefined
\let\noarapdir\undefined
\let\nocandir\undefined
\let\methodfig\undefined
\let\imfig\undefined
\let\rainbowours\undefined
\let\methtext\undefined
\let\comparerow\undefined
\let\linegap\undefined
\let\cdpofig\undefined
\let\imfig\undefined
\let\rainbowours\undefined

\graphicspath{{images/qualitative/}}
\newcommand{\cdpodir}{dpc3dpo-h36m21_017_c3dpo_00000_MOD.camera_scale=True_M.l.l_canonicalization=0.1}
\newcommand{\oursdir}{dpc3dpo-h36m21_021_weight_repro_00003_M.l.l_arap=0.3_MOD.shared_scale=False}
\newcommand{\lbocdpodir}{dpc3dpo-h36m21_018_lboc3dpo_00000_MOD.lbo_vertex_transform_model=translation_M.l.l_canonicalization=1.0_MOD.shape_basis_size=10}
\newcommand{\noarapdir}{dpc3dpo-h36m21_025_reprow_ablation_00003_M.l.l_canonicalization=0.1_M.l.l_part_entropy=0.001_M.l.l_arap=0.0_MOD.shared_scale=False}
\newcommand{\noentrdir}{dpc3dpo-h36m21_025_reprow_ablation_00005_M.l.l_canonicalization=0.1_M.l.l_part_entropy=0.0_M.l.l_arap=0.3_MOD.shared_scale=False}
\newcommand{\nocandir}{dpc3dpo-h36m21_025_reprow_ablation_00009_M.l.l_canonicalization=0.0_M.l.l_part_entropy=0.001_M.l.l_arap=0.3_MOD.shared_scale=False}

\newcommand{\imh}{1.72cm}
\newcommand{\linegap}{\vspace{-0.0cm}}

\newcommand{\methodfig}[2]{%
\includegraphics[trim=1cm 0.25cm 1cm 1.6cm, clip, height=\imh]{#1/#2.png}%
}

\newcommand{\cdpofig}[2]{%
\includegraphics[trim=0.5cm 0cm 0.5cm 0.85cm, clip, height=\imh]{#1/#2.png}%
}

\newcommand{\imfig}[1]{%
\includegraphics[height=2.5cm]{\oursdir/#1_image_kp_partseg.png}%
}

\newcommand{\rainbowours}{%
{\color[HTML]{8D274B}D}%
{\color[HTML]{A48053}P}%
{\color[HTML]{547CA2}3}%
{\color[HTML]{8EA395}D}%
}

\newcommand{\methtext}[1]{\centering\scriptsize\hspace{0.01cm}\rotatebox{90}{\hspace{0.1cm}#1}\hspace{0.01cm}}

\newcommand{\comparerow}[1]{%
\begin{minipage}[c]{0.24\linewidth}\centering%
    \imfig{#1}\\\linegap
    \methtext{~\textbf{\color[HTML]{960061}C3DPO \cite{Novotny2019}}}%
    \cdpofig{\cdpodir}{#1_pcl_lbo_filtered_solid_000_000}%
    \cdpofig{\cdpodir}{#1_pcl_lbo_filtered_solid_000_089}\\\linegap
    \methtext{~~~w/o ARAP}%
    \methodfig{\noarapdir}{#1_mesh_raw_partseg_000_000}%
    \methodfig{\noarapdir}{#1_mesh_raw_partseg_000_089}\\\linegap
    \methtext{w/o canonical'n}%
    \methodfig{\nocandir}{#1_mesh_raw_partseg_000_000}%
    \methodfig{\nocandir}{#1_mesh_raw_partseg_000_089}\\\linegap
    \methtext{~~w/o entropy}%
    \methodfig{\noentrdir}{#1_mesh_raw_partseg_000_000}%
    \methodfig{\noentrdir}{#1_mesh_raw_partseg_000_089}\\\linegap
    \methtext{~~~~w/o parts}%
    \methodfig{\lbocdpodir}{#1_mesh_raw_solid_000_000}%
    \methodfig{\lbocdpodir}{#1_mesh_raw_solid_000_089}\\\linegap
    \methtext{~\textbf{\rainbowours~\color[HTML]{000000}(ours)}}%
    \methodfig{\oursdir}{#1_mesh_raw_partseg_000_000}%
    \methodfig{\oursdir}{#1_mesh_raw_partseg_000_089}%
\end{minipage}}

\begin{figure*}
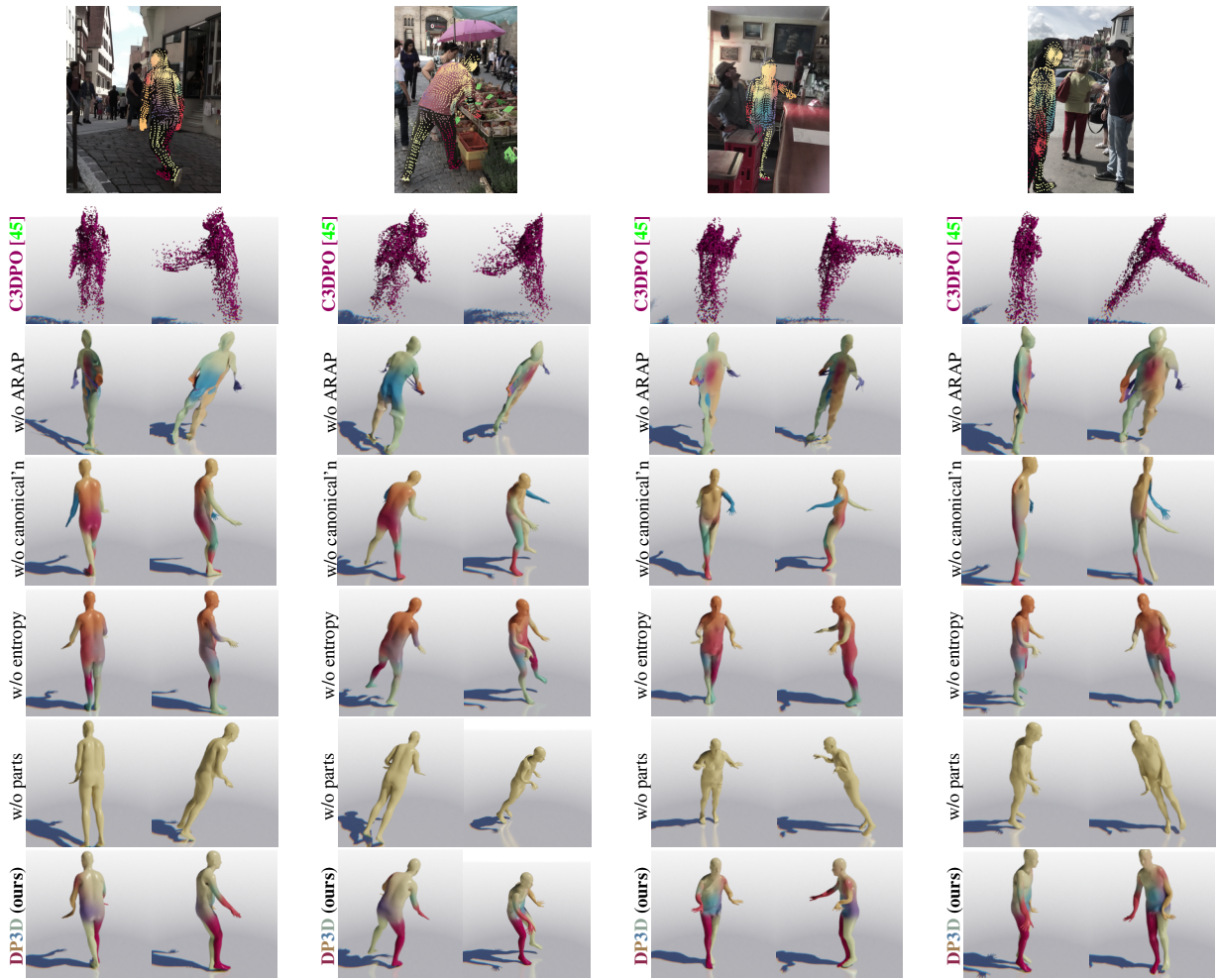


\comparerow{downtown_walkUphill_00/image_00312}
\comparerow{downtown_weeklyMarket_00/image_00532}
\comparerow{downtown_bar_00/image_00793}
\comparerow{downtown_bus_00/image_00485}

\caption{\textbf{Qualitative evaluation on 3DPW}.
From top to bottom: input image and keypoints,
the following rows show the reconstruction of C3DPO~\cite{Novotny2019}, the results of DP3D without corresponding losses, with the linear model instead of parts segmentation,
and of the full proposed method.%
}\label{f:3dpwres-sup}
\end{figure*}

\let\cdpodir\undefined
\let\lbocdpodir\undefined
\let\oursdir\undefined
\let\imh\undefined
\let\noentrdir\undefined
\let\noarapdir\undefined
\let\nocandir\undefined
\let\methodfig\undefined
\let\imfig\undefined
\let\rainbowours\undefined
\let\methtext\undefined
\let\comparerow\undefined
\let\linegap\undefined
\let\cdpofig\undefined
\let\imfig\undefined
\let\rainbowours\undefined

\graphicspath{{images/qualitative/}}
\newcommand{\lbocdpodir}{dpc3dpo-standogs_007_lboc3dpo_00005_MOD.shared_scale=True_M.l.l_canonicalization=0.1_MOD.lbo_num_harmonics=64_MOD.weight_reprojection_by_area=True}
\newcommand{\oursdir}{dpc3dpo-standogs_010_ablation_00033_MOD.shared_scale=True_M.l.l_canonicalization=0.1_M.l.l_part_entropy=0.001_MOD.lbo_num_harmonics=64_MOD.num_latent_parts=10_MOD.shape_basis_size=5}
\newcommand{\noarapdir}{dpc3dpo-standogs_011_ablation_00057_M.l.l_canonicalization=0.0_M.l.l_part_entropy=0.001_M.l.l_local_rigidity=0.0_M.l.l_arap=0.0_MOD.lbo_num_harmonics=64_MOD.shape_basis_size=5}
\newcommand{\noentrdir}{dpc3dpo-standogs_010_ablation_00041_MOD.shared_scale=True_M.l.l_canonicalization=0.1_M.l.l_part_entropy=0.0_MOD.lbo_num_harmonics=64_MOD.num_latent_parts=10_MOD.shape_basis_size=5}
\newcommand{\nocandir}{dpc3dpo-standogs_010_ablation_00049_MOD.shared_scale=True_M.l.l_canonicalization=0.0_M.l.l_part_entropy=0.001_MOD.lbo_num_harmonics=64_MOD.num_latent_parts=10_MOD.shape_basis_size=5}

\newcommand{\imh}{1.8cm}

\newcommand{\methodfig}[2]{%
\includegraphics[trim=1cm 0.5cm 1cm 1cm, clip, height=\imh]{#1/val_#2.png}%
}

\newcommand{\imfig}[1]{%
\includegraphics[height=\imh]{\oursdir/val_#1_image_kp_partseg}%
}

\newcommand{\rainbowours}{%
{\color[HTML]{8D274B}D}%
{\color[HTML]{A48053}P}%
{\color[HTML]{547CA2}3}%
{\color[HTML]{8EA395}D}%
}

\newcommand{\methtext}[1]{\centering\scriptsize\hspace{0.01cm}\rotatebox{90}{\hspace{0.01cm}#1}\hspace{0.01cm}}

\newcommand{\comparerow}[1]{%
\begin{minipage}[c]{0.24\linewidth}\centering%
    \imfig{#1}\\
    \methtext{~~~w/o ARAP}%
    \methodfig{\noarapdir}{#1_mesh_raw_partseg_000_000}%
    \methodfig{\noarapdir}{#1_mesh_raw_partseg_000_089}\\
    \methtext{w/o canonical'n}%
    \methodfig{\nocandir}{#1_mesh_raw_partseg_000_000}%
    \methodfig{\nocandir}{#1_mesh_raw_partseg_000_089}\\
    \methtext{~~w/o entropy}%
    \methodfig{\noentrdir}{#1_mesh_raw_partseg_000_000}%
    \methodfig{\noentrdir}{#1_mesh_raw_partseg_000_089}\\
    \methtext{~~~~w/o parts}%
    \methodfig{\lbocdpodir}{#1_mesh_raw_solid_000_000}%
    \methodfig{\lbocdpodir}{#1_mesh_raw_solid_000_089}\\%
    \methtext{~~~~\textbf{\rainbowours~\color[HTML]{000000}(ours)}}%
    \methodfig{\oursdir}{#1_mesh_raw_partseg_000_000}%
    \methodfig{\oursdir}{#1_mesh_raw_partseg_000_089}\\%
    \methtext{~~GT SMAL fits}
    \methodfig{\oursdir}{#1_gt_mesh_raw_err_000_000}
    \methodfig{\oursdir}{#1_gt_mesh_raw_err_000_089}
\end{minipage}}

\begin{figure*}
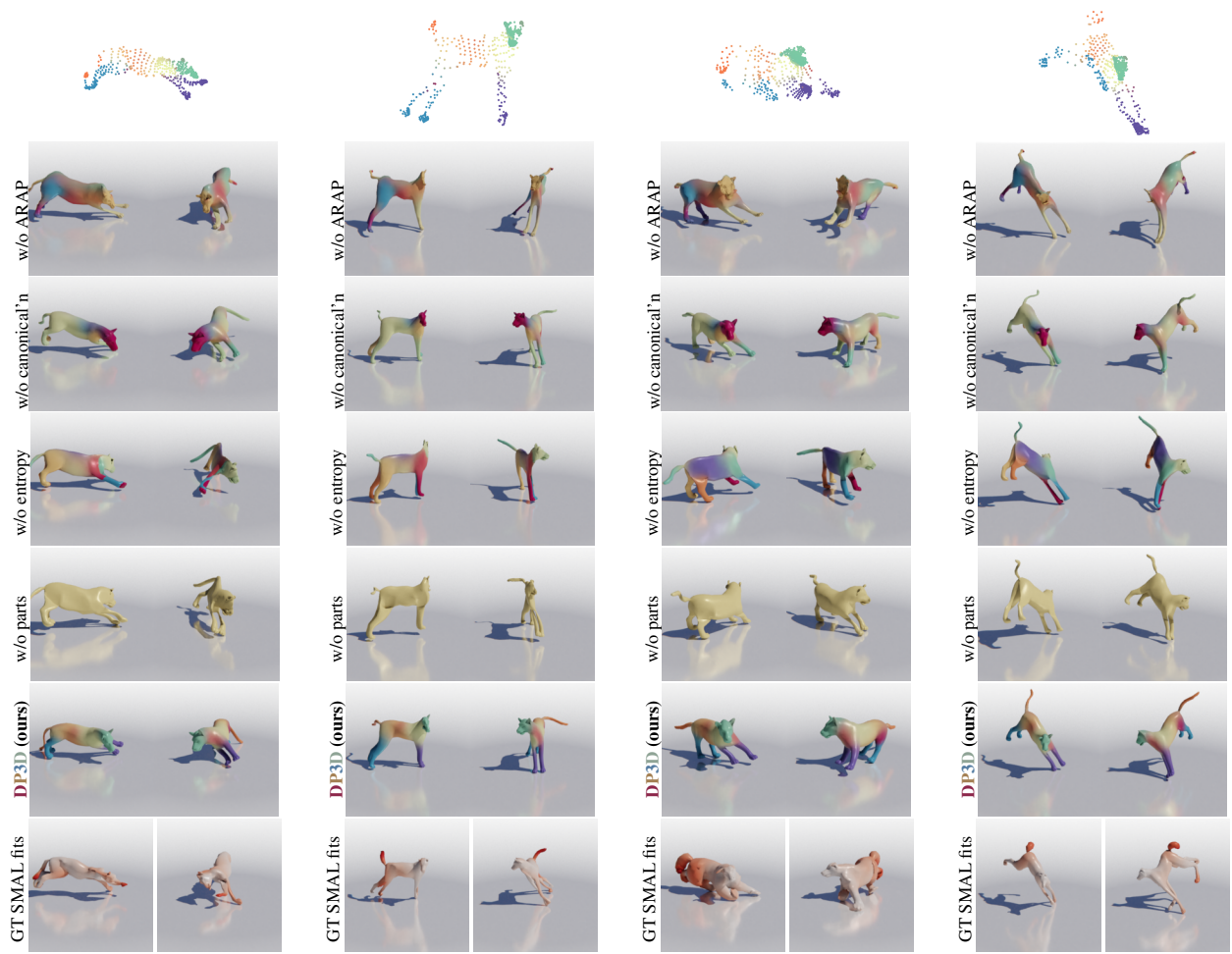

\comparerow{00000899}
\comparerow{00003917}
\comparerow{00000120}
\comparerow{00003909}

\caption{\textbf{Results on Stanford Dogs}. 
The first row shows input keypoints obtained by projecting SMPL fits showed in the last row,
the following rows show the results of DP3D without corresponding losses, of the no-parts baseline and of our reconstruction from the camera's and from an alternative viewpoint,
the last row color-codes errors on the ``ground-truth'' mesh.%
}\label{f:dogsres-sup}
\end{figure*}

\let\cdpodir\undefined
\let\lbocdpodir\undefined
\let\oursdir\undefined
\let\imh\undefined
\let\noentrdir\undefined
\let\noarapdir\undefined
\let\nocandir\undefined
\let\methodfig\undefined
\let\imfig\undefined
\let\rainbowours\undefined
\let\methtext\undefined
\let\comparerow\undefined
\let\linegap\undefined
\let\cdpofig\undefined
\let\imfig\undefined
\let\rainbowours\undefined

\fi

{\small\bibliographystyle{ieee_fullname}\bibliography{egbib,vedaldi_general,vedaldi_specific}}

\end{document}